\RequirePackage{fix-cm}
\documentclass[twocolumn]{svjour3}          
\smartqed  
\usepackage{subcaption}
\usepackage{caption}
\usepackage{times}
\usepackage{graphicx}
\usepackage{amsmath}
\usepackage{amssymb}
\usepackage{booktabs}
\usepackage{makecell}
\usepackage{multirow}
\usepackage{colortbl}
\usepackage{tabu}
\usepackage{xspace}
\usepackage{pifont}
\usepackage{enumitem}
\newcommand{\cmark}{\ding{51}\xspace}%
\newcommand{\xmarkg}{\textcolor{lightgray}{\ding{55}}\xspace}%

\usepackage{epsfig}
\usepackage[]{footmisc}
\usepackage{pdfpages}

\usepackage[pagebackref=true,breaklinks=true,letterpaper=true,colorlinks,bookmarks=false]{hyperref}

\newcommand{\ntacc}{N-acc.\xspace}
\newcommand{\tacc}{T-acc.\xspace}

\let\oldsubsection\subsection
\renewcommand{\subsection}[1]{\oldsubsection{#1} }

\usepackage{color}
\usepackage{xcolor}

\usepackage{mathrsfs}
\usepackage{soul}
\newcommand{\ie}{{\emph{i.e.}}\xspace}
\newcommand{\eg}{{\emph{e.g.}}\xspace}
\newcommand{\Eg}{{\emph{E.g.}}\xspace}
\newcommand{\vs}{{\emph{v.s.}}\xspace}
\newcommand{\etal}{{\emph{et al.}}\xspace}

\usepackage{array}
\usepackage{tabu}

\definecolor{Gray}{gray}{0.9}
\definecolor{pink}{rgb}{1.0, 0.85, 0.85}

\journalname{International Journal of Computer Vision}

\hypersetup{pdfstartview={FitH}}
\hypersetup{pdfborder={0 0 0}}
\hypersetup{pdfpagemode={UseNone}}
\begin{document}

\title{GREx: Generalized Referring Expression Segmentation, Comprehension, and Generation}

\author{Henghui Ding \and Chang Liu \and Shuting He \and Xudong Jiang \and Yu-Gang Jiang
}

\institute{Henghui Ding ({henghui.ding@gmail.com}), Fudan University, China.
}

\date{}
\maketitle

\begin{abstract}

Referring Expression Segmentation (RES) and Comprehension (REC) respectively segment and detect the object described by an expression, while Referring Expression Generation (REG) generates an expression for the selected object. Existing datasets and methods commonly support single-target expressions only, \ie, one expression refers to one object, not considering multi-target and no-target expressions. This greatly limits the real applications of REx (RES/REC/REG). This paper introduces three new benchmarks called Generalized Referring Expression Segmentation (GRES), Comprehension (GREC), and Generation (GREG), {collectively denoted as GREx}, which extend the classic REx to allow expressions to identify an arbitrary number of objects. We construct the first large-scale GREx dataset gRefCOCO that contains multi-target, no-target, and single-target expressions and their corresponding images with labeled targets. {GREx and gRefCOCO are designed to be backward-compatible with REx, facilitating extensive experiments to study the performance gap of the existing REx methods on GREx tasks}. One of the challenges of GRES/GREC is complex relationship modeling, for which we propose a baseline ReLA that adaptively divides the image into regions with sub-instance clues and explicitly models the region-region and region-language dependencies. The proposed ReLA achieves the state-of-the-art results on the both GRES and GREC tasks. The proposed gRefCOCO dataset and method are available at \href{https://henghuiding.com/GREx}{https://henghuiding.com/GREx}.
\end{abstract}

\begin{figure*}
    \centering
    \includegraphics[width=1\textwidth]{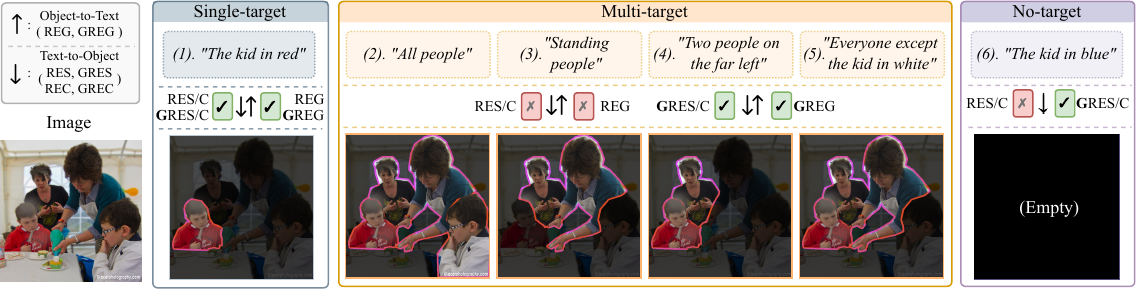} 
\vspace{-6mm}
\captionof{figure}{{Classic Referring Expression Segmentation (RES), Comprehension (REC), and Generation (REG), collectively denoted as REx, only supports expressions that indicate a single target object, \eg, \textit{``The kid in red''}. Compared with REx, the proposed \textbf{Generalized Referring Expression tasks (GREx), including Generalized RES (GRES), Generalized REC (GREC), and Generalized REG (GREG)}, extend expressions to multi-target or no-target. For example, GREx support multi-target expressions that indicate several objects by their commonalities or relationships, \eg, category \textit{(2) ``All {people}''}, attribute \textit{(3) ``Standing people''}, counting \textit{(4) ``Two people on the far left''}, and compound \textit{(5) ``Everyone except the kid in white''}. GRES and GREC further support no-target expressions that do not match any object, \eg, \textit{(6)~``The kid in blue''}.}}
    \label{fig:fig1} 
\end{figure*}

\section{Introduction}\label{sec:intro}

Referring Expression Segmentation (RES), Referring Expression Comprehension (REC), and Referring Expression Generation (REG) represent three significant and emerging tasks in the field of multi-modal information processing~\cite{ding2025multimodal}. These tasks inherently bridge the domains of computer vision and natural language processing, showing their growing importance. When provided with an image and a natural language expression describing an object in that image, both RES and REC tasks are focused on locating the specified target object. RES aims to predict a segmentation mask for the target object, while the gole of REC is to predict a bounding box. {In contrast to RES and REC that focus on understanding the given referring expression and grounding the corresponding target object, REG is a generative task that aims to generate an unambiguous referring expression for the target object selected by a bounding box in the given image.} The applicability of RES, REC, and REG spans various domains, \eg, image editing, caption, video production, human-machine interaction, enabling a diverse range of practical applications. Currently, most of the existing methods in the field of referring expression adhere to the default rules defined in the influential datasets ReferIt \cite{kazemzadeh-etal-2014-referitgame} and RefCOCO \cite{yu2016modeling,mao2016generation}. These default rules govern the quantity and nature of expressions and their corresponding targets. Under this paradigm, previous methods~\cite{VLTPAMI,yu2018mattnet,yu2017joint} have experienced notable advancements over recent years, showcasing their effectiveness in {understanding or generating} single-target expressions that refer to one object.

\textbf{Limitations of classic RES, REC, {and REG}.} However, most classic RES, REC, and REG methods are bound by inherent limitations stemming from their pre-defined constraints. First, these methods do not account for scenarios wherein the referring expression does not align with any object present in the given image. Consequently, the response of the established RES and REC methods remains undefined in such situations. When it comes to practical applications under such a constraint, the input expression must be precisely linked to a particular object in the image, otherwise, problems caused by incorrect predictions are bound to arise. Second, most existing referring expression datasets, such as the widely-used RefCOCO \cite{yu2016modeling,mao2016generation}, do not contain multi-target expressions that refer to multiple objects. {For REG task, this limitation neglects the need to describe multiple objects with a single sentence in real-world scenarios.} For RES and REC, this limitation compels the requirement of multiple sequential expression inputs to separately identify objects one after another within an image. As shown in \figurename~\ref{fig:fig1}, segmenting \textit{``All people''} requires four separate expressions, resulting in four model calls. {Although open-vocabulary segmentation and detection~\cite{OpenVocabulary} can return all instances of a given category name like \textit{``people''}, they cannot handle free-form expressions that target selective subsets of same-category instances or involve attributes, relations, or other cues, \eg, \textit{``Everyone except the kid in white''}.} Our experiments demonstrate that the classic RES, REC, and REG methods, trained on existing datasets and predefined constraints, are insufficient in achieving generalization across these complex diverse scenarios.

\textbf{New GREx benchmarks and gRefCOCO dataset.} In this work, in order to overcome the limitations of {classic RES, REC, and REG, we introduce three new GREx benchmarks, called Generalized Referring Expression Segmentation (GRES), Generalized Referring Expression Comprehension (GREC), and Generalized Referring Expression Generation (GREG)}, which allow expressions indicating any number of target objects. GRES/GREC takes an image and a referring expression as input, the same as classic RES/REC. As shown in \figurename~\ref{fig:fig1}, in contrast to the classic RES and REC, which focus on single-target expressions, GRES and GREC further support multi-target expressions that refer to multiple target objects of a given image in a single expression, \eg, \textit{``Everyone except the kid in white''} in \figurename~\ref{fig:fig1}, and no-target expressions that do not correspond to any object within the image, \eg, \textit{``the kid in blue''} in \figurename~\ref{fig:fig1}. {Compared to classic REG focusing on single object, GREG additionally supports describing a set of multiple selected objects unambiguously and naturally with a single sentence.} By allowing expressions to refer to any number of target objects, GREx introduce a heightened level of flexibility in inputs. This expanded capability allows for a more natural, user-friendly, and intuitive way of language interacting with images, which significantly enhances the usefulness and robustness of referring expression perception and generation in practical applications. Previous referring expression datasets~\cite{kazemzadeh-etal-2014-referitgame,yu2016modeling,mao2016generation} have not been designed to include samples featuring multi-target expressions or no-target expressions. These datasets predominantly comprise single-target expressions, as outlined in \tablename~\ref{tab:dataset_compare}. This underscores the requirement for more comprehensive datasets that can more accurately reflect the real-world scenarios. {To support research efforts towards more realistic and practical referring expression understanding and generation, we build a new dataset for GREx, termed gRefCOCO}. This dataset is an extension of the well-known RefCOCO~\cite{yu2016modeling,mao2016generation} and introduces two distinctive sample types that are absent from existing datasets: 1) multi-target samples, wherein the expression refers to two or more target objects in the given image, and 2) no-target samples, wherein the expression fails to match any object in the image.
Following the introduction of the GRES task~\cite{GRES}, several complementary datasets~\cite{beyond1to1,wu2023advancing} have emerged, highlighting the growing attention to GRES. For example, Ref-ZOM~\cite{beyond1to1} supports multi- and no-target expressions, but many are synthetically composed or randomly paired with captions. GRD~\cite{wu2023advancing} adopts cross-image group retrieval but includes only 316 expressions with limited diversity, see \tablename~\ref{tab:dataset_compare}. In contrast, gRefCOCO systematically support GREx with rich, realistic expressions grounded in instance masks and boxes, providing a more comprehensive benchmark.

\begin{table}[t]
  \renewcommand\arraystretch{1.2}
  \centering
  \footnotesize
  \caption{{Comparison among referring expression datasets, including ReferIt~\cite{kazemzadeh-etal-2014-referitgame}, RefCOCO(g)~\cite{yu2016modeling,mao2016generation}, PhraseCut~\cite{wu2020phrasecut}, Ref-ZOM~\cite{beyond1to1}, GRD~\cite{wu2023advancing}, and the proposed \textbf{gRefCOCO}. $\mathcal{S}$: single-target expression that refers to a single target object in the image. $\mathcal{M}$: multi-target expression that refers to multiple target objects in the image. $\mathcal{N}$: no-target expression that fails to correspond to any object in the image. $\mathbb{M}$: Mask annotation. $\mathbb{B}$: Bounding box annotation.}}
  \vspace{-2.6mm}
  \setlength{\tabcolsep}{0.36mm}
  \resizebox{0.48\textwidth}{!}{{\begin{tabular}{lcccccc}
    \specialrule{.1em}{.05em}{.05em} 
          & ReferIt & RefCOCO(g)& PhraseCut&Ref-ZOM & GRD&  \textbf{gRefCOCO}\\
          \cline{2-7}
          Source  & CLEF~\cite{grubinger2006iapr} & COCO~\cite{lin2014microsoft} & VG~\cite{krishna2017visual}  & COCO&Internet & COCO \\
          $\mathcal{S}$-target& {\cmark} & {\cmark} & {\cmark} & {\cmark}& {\cmark}& {\cmark} \\
          $\mathcal{M}$-target & {\xmarkg} & {\xmarkg} & {fallback} & {\cmark}& {\cmark}& {\cmark} \\
          $\mathcal{N}$-target & {\xmarkg} & {\xmarkg} & {\xmarkg}  & {\cmark}& {\cmark}& {\cmark} \\
        \#Expr.& 120k& 142k/95k&  345k& 90k& 0.3k& 259k \\
         Annot. & $\mathbb{M}\&\mathbb{B}$  &  $\mathbb{M}\&\mathbb{B}$  &  $\mathbb{M}\&\mathbb{B}$ & $\mathbb{M}\&\mathbb{B}$ &  $\mathbb{M}$&  $\mathbb{M}\&\mathbb{B}$ \\
          Expr. type & free  & free  & templated &free & group& free \\
    \specialrule{.1em}{.05em}{.05em} 
    \end{tabular}}%
    }
  \label{tab:dataset_compare}%
\vspace{-2.36mm}
\end{table}%

\textbf{A baseline method for GRES and GREC.} Furthermore, we propose a baseline method for GRES and GREC. It is widely recognized that the inclusion of relationship modeling, such as interactions between regions, is pivotal for successful RES and REC \cite{yu2018mattnet}. Nonetheless, classic RES and REC methods typically focus on detecting single object, allowing some methods to achieve satisfactory performance without explicit region-to-region interaction modeling. However, in the context of GRES and GREC, where multi-target expressions involve multiple objects in a single expression, the intricacy of modeling long-range region-to-region dependencies becomes more pronounced and imperative. Hence, taking this aspect into consideration, we propose a region-based approach, ReLA, tailored for GRES and GREC. This method splits the image into semantic regions and explicitly models the interaction among these regions with sub-instance clues, enabling a more nuanced capture of the interactions. In contrast to previous methods where regions originate from a straightforward and hard-split of the input image, ReLA employs a soft-aggregation strategy to compile features for individual regions, creating enhanced flexibility into the process. Compared to our previous work~\cite{GRES}, an enhanced ReLA is proposed in this journal version to learn GRES and GREC simultaneously in a unified framework. We conduct comprehensive experiments on our proposed methods in comparison with existing RES and REC methods. Our results showcase the significant impact of explicitly modeling interactions and extracting features from flexibly soft-divided regions on the performance of GRES and GREC tasks.

In summary, our contributions are listed as follows:
\vspace{-1mm}
\begin{itemize}
  \item[1)] {We propose three new GREx benchmarks: Generalized Referring Expression Segmentation (GRES), Generalized Referring Expression Comprehension (GREC), and Generalized Referring Expression Generation (GREG), making RES, REC, and REG flexible and practical in real-world scenarios.}
  
  \vspace{0.36mm}
  \item[2)] We create a large-scale dataset, named as gRefCOCO, to facilitate the future research in exploring generalized referring expression segmentation, comprehension, and generation. To the best of our knowledge, the introduced gRefCOCO dataset pioneers the support for expressions that refer to an arbitrary number of target objects.
  
  \vspace{0.36mm}
  \item[3)] To capture fine-grained sub-instance attributes and model complex \textbf{ReLA}tionships among objects, we propose a baseline method ReLA for GRES and GREC. 
  
  \vspace{0.36mm}
  \item[4)] {By defining evaluation metrics and conducting comprehensive experiments, we closely examine the newly introduced GREx tasks along with the gRefCOCO dataset. We analyze the emerging challenges intrinsic to GREx and provide potential directions for future research.} 
\end{itemize}

\vspace{-3mm}
\section{Related Works} \label{sec:related_works}
\textbf{Related referring tasks and datasets.} Referring Expression \textit{Comprehension} (REC) \cite{hu2016natural} aims to predict a bounding box for the target object in the input image that is described by the given expression, while Referring Expression \textit{Segmentation} (RES) \cite{hu2016segmentation} aims to predict a segmentation mask for the target object. The earliest dataset for RES and REC is ReferIt \cite{kazemzadeh-etal-2014-referitgame}. However, ReferIt is not initially designed for RES and REC but for Referring Expression \textit{Generation} (REG)~\cite{kazemzadeh-etal-2014-referitgame}, which aims to generate an expression for a selected segment. Thus, one ReferIt expression can only refer to one segment. {Although ReferIt~\cite{kazemzadeh-etal-2014-referitgame} has a small number (less than 5\%) of expressions that refer to multiple objects, all of them are restricted in a same region of image, inherited from its base dataset SAIAPR~\cite{SAIAPR}, which is not strictly instance nor semantic level but {a little} haphazardly segments image into several ``regions''. ReferIt gives one expression to one such ``region'' that sometimes covers multiple objects. So it cannot provide individual instance-level masks like gRefCOCO. Moreover, these multi-objects are not intentionally selected by some meaning but are just located together.} Later on, RefCOCO and RefCOCO+ datasets are introduced by Yu \etal \cite{yu2016modeling} to support RES and REC. Nevertheless, RefCOCO is confined to single-target expressions. A similar well-known dataset, RefCOCOg \cite{mao2016generation}, also adheres to this limitation. REC is typically defined as the task of grounding a single target object in an input image using a given referring expression~\cite{hu2016natural}. Although the original definition of RES \cite{hu2016segmentation} does not limit the number of target instances, ``\textbf{\textit{one expression, one object}}'' has become a ``de-facto'' rule for both RES and REC tasks. Furthermore, it's important to note that, to the best of our knowledge, all previous methods and datasets do not support expressions that miss all targets in the image and refer to some targets not existing in the image, \ie, no-target expressions.

In recent years, several new datasets have emerged. However, most of them neither emphasize nor align well with the GREx tasks. For example, PhraseCut dataset \cite{wu2020phrasecut} includes some multi-target expressions, but only as ``fallback'' options when an object cannot be uniquely referred to. Furthermore, expressions in PhraseCut are constructed using templates, limiting the sentence diversity. 
Datasets for image captioning, such as Flickr30K \cite{plummer2015flickr30k} and Visual Genome \cite{krishna2017visual}, share similarities with REx. However, it's worth noting that the expressions in these datasets are centered around describing the given image/object, rather than distinguishing between different instances. Consequently, they do not inherently guarantee the disambiguation of expression$\rightarrow$object(s) and are not feasible for referring expression tasks. While there are referring datasets that leverage alternative data modalities or learning schemes, like ScanRefer \cite{chen2020scanrefer} which focuses on 3D objects, and ClevrTex \cite{karazija2021clevrtex} which centers on unsupervised learning, they do not support expressions in indicating multiple target objects. Moreover, none of the previously mentioned datasets incorporate no-target expressions. 
Following the introduction of GRES~\cite{GRES}, there are several new datasets~\cite{beyond1to1,wu2023advancing} focusing on complementary aspects emerged in the subsequent conferences. For example, Ref-ZOM~\cite{beyond1to1} includes multi-target and no-target expressions. However, many of its multi-target samples are synthetically composed by merging single-target expressions or using category templates, while its no-target cases are created by randomly pairing images with unrelated captions. GRD~\cite{wu2023advancing} supports multi- and no-target scenarios through cross-image group retrieval, but contains only 316 expressions with limited diversity. In contrast, gRefCOCO is the first to systematically define GREx with rich, realistic expressions grounded on instance masks and boxes, offering a more comprehensive and well-defined benchmark. Together, these works underscore the growing trend and popularity of GREx~\cite{GRES}.

\textbf{Referring expression segmentation (RES) methods.} RES methods can be broadly categorized into two main groups: one-stage (or top-down) methods~\cite{margffoy2018dynamic,zhang2022coupalign,ding2020phraseclick,li2018referring,chen2019see,ye2019cross,hu2020bi,huang2020referring,hui2020linguistic,luo2020cascade} and two-stage (or bottom-up) methods~\cite{yu2018mattnet,liu2022instance,Chen_lang2seg_2019}. One-stage methods have an FCN-like \cite{long2015fully} end-to-end network, and the prediction is achieved by per-pixel classification on fused multi-modal feature. 
Representative works include LTS~\cite{jing2021locate} and ISFP~\cite{liu2022instance} that first give a rough location of the target object and then produce the target mask, and MCN~\cite{luo2020multi} that combine bounding boxes in RES and segmentation masks in REC together. 
Two-stage methods, \eg, MattNet~\cite{yu2018mattnet}, first employ a pre-existing instance segmentation network to generate a set of instance proposals. Subsequently, they determine the target by selecting from among these generated proposals. Ding \etal~\cite{VLTPAMI,ding2021vision} introduce transformer~\cite{carion2020end} into RES and propose Vision-Language Transformer (VLT) to deal with vision and language tokens. After that, more transformer-based methods~\cite{yang2021lavt,wang2022cris,CGFormer,PolyFormer,GRES,UNINEXT,xu2023bridging} are proposed and bring large performance gains compared to CNN-based methods. 

Since the introduction of the GRES task~\cite{GRES}, an increasing number of methods \cite{li2024bring,GSVA,LQMFormer,LuoIJCV,zhang2024psalm,HieA2G,LISA} are proposed to address this challenge. For example, MABP~\cite{li2024bring} introduces adaptive binding of queries to regional object features, enabling flexible matching for multi-target and no-target expressions in GRES while easing encoder-decoder coupling. GSVA~\cite{GSVA} uses MLLMs and introduces specialized [\texttt{SEG}] tokens for multi-target cases, along with a [\texttt{REJ}] token to explicitly reject irrelevant queries in no-target cases.

\textbf{Referring expression comprehension (REC) methods.} REC predicts a bounding box for the target object~\cite{hu2016natural,wang2019neighbourhood,liu2019learning,yang2019fast,zhuang2018parallel,yang2020improving,liao2020real,RefCLIP}. Earlier REC works typically use a multi-stage pipeline \cite{liu2019learning,zhuang2018parallel,hu2017modeling,zhang2018grounding,hong2019learning}, which utilizes a pre-trained object detection network~\cite{he2017mask} to generate a collection of instance proposals for the input image. The proposals are then compared against the given language expression to identify the most suitable match. One example of a two-stage method is MAttNet by Yu \etal~\cite{yu2018mattnet}. MAttNet leverages Mask R-CNN~\cite{he2017mask} to detect all instances in the image in the first stage, and a modular network is then used in the second stage to match and select the target object from the detected instances. Nevertheless, two-stage methods have high computational costs, and their performance depends on the first stage detection network. To reach real-time processes and better grounding performance, there has been a growing trend towards using one-stage methods in recent years, such as \cite{chen2018real,liao2020real,yang2019fast,sun2022proposal,deng2021transvg}. For example, Yang \etal~\cite{yang2019fast} concatenates text embedding into the visual feature of real-time detector YOLOv3~\cite{yolov3}. Transformer-based methods~\cite{deng2021transvg,MDETR,GroundingDINO} recently demonstrate powerful improvement. For example, TransVG~\cite{deng2021transvg} employs visual branch and linguistic branch to extract visual and linguistic tokens, respectively, and inputs these tokens to a visual-linguistic transformer. MDETR~\cite{deng2021transvg} detects the target object(s) using text query as conditional tokens. {GroundingDINO~\cite{GroundingDINO} is widely adopted for its grounding accuracy and streamlined transformer-based framework. Building upon it, MM-Grounding-DINO~\cite{zhao2024open} improves performance by introducing more deliberately designed training strategies. Large language model (LLM) pipelines are another emerging direction. LLM-wrapper~\cite{LLM-wrapper} uses a frozen black-box VLM (\eg, GroundingDINO) to generate candidate boxes, then employs an LLM to match them with the referring expression and select the best-matched one. Shikra~\cite{Shikra} proposes a unified framework that treats spatial coordinates as natural language, enabling bidirectional grounding and captioning.}

\textbf{Referring expression generation (REG) methods.} REG aims to generate an unambiguous natural language expression given an image and a bounding box indicating an object in this image. Though it can be seen as an inverse task to Referring Expression Comprehension (REC), it is one of the traditional tasks for natural language generation, which can be traced back to 1990s~\cite{reiter1997building}. In the past decade, as the emerge of deep learning, REG has been greatly advanced and many fundamental works are proposed, \eg, the first large-scale dataset RefCLEF~\cite{rohrbach2016grounding}, and RefCOCO family datasets~\cite{kazemzadeh-etal-2014-referitgame,mao2016generation,yu2016modeling}. Many works are proposed to enhance the usage of features and the generation quality~\cite{kim2020conan,tanaka2019generating,ye2023whether,schuz2021decoupling}. Yu~\etal proposes a ``speaker - listener'' pipeline~\cite{yu2017joint} to jointly train REG and REC together.
{Recent advances in multi-modal pretraining have inspired many methods to generate referring expressions directly from images using large-scale vision-language models or LLMs~\cite{bracha2023disclip,xiaoke2023SCA,yang2024enhancing}.
For example, Liang et al.~\cite{liang2024unleashing} propose a training-free framework, unleash-then-eliminate, which extracts latent cues from intermediate layers and applies a cycle-consistency decoding step to reduce hallucinations in the REG task. In addition, generalist generative models such as \cite{yu2023merlin,peng2023kosmos} have shown the ability to perform REG. However, similar to classic RES and REC methods, these approaches remain limited to generating expressions for a single target object.}

{\textbf{Referring Expression Multi-Task Methods.} Multi-task learning has become a common paradigm in segmentation, detection, and generation, where a shared backbone is combined with lightweight, task-specific heads. Following this way, several works~\cite{mcn,segvg} address REC and RES jointly, while others~\cite{lang2seg,yang2024enhancing} extend the collaboration to include REG. Recent advances in multimodal large language models have further driven the pursuit of unified frameworks. For example, GLaMM~\cite{GLaMM} generates natural language descriptions along with corresponding segmentation masks, handling region-level captioning and RES simultaneously. Florence-2~\cite{Florence-2} advances this concept at foundation-model scale by offering a prompt-based interface that supports REC, RES, captioning, and other vision-language tasks within a single framework.}

\section{Task Setting and Dataset}

\subsection{GREx Task Settings}
\vspace{-2mm}
\textbf{Revisiting Classic RES and REC.} Classic Referring Expression Segmentation (RES) and Referring Expression Comprehension (REC) take an image and an expression as inputs. The objective is to generate a segmentation mask for RES or a bounding box for REC corresponding to the object indicated by the input expression. As mentioned in Sec.~\ref{sec:related_works}, previous RES and REC datasets as well as methods do not account for no-target expressions. Moreover, all samples in existing RES and REC datasets predominantly pertain to single-target expressions. Consequently, current methods are inclined to produce erroneous outputs if the input expression refers to either nothing or multiple target objects within the given image.

\begin{figure}[t]
  \centering
  \begin{subfigure}[t]{\linewidth}
      \centering
      \includegraphics[width=\textwidth]{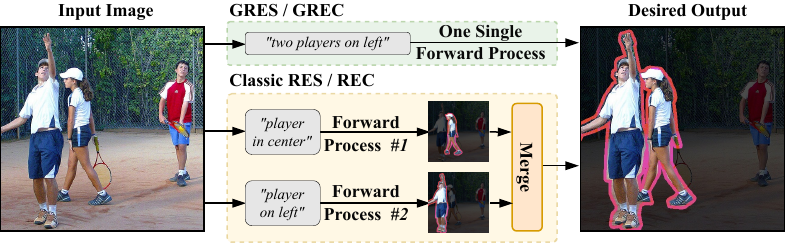}
      \vspace{-5mm}
      \caption{Multi-target: selecting multiple objects in a single forward process.}
      \label{fig:app_mt}
      \vspace{2mm}
  \end{subfigure} \\
  \begin{subfigure}[t]{\linewidth}
      \centering
      \includegraphics[width=\textwidth]{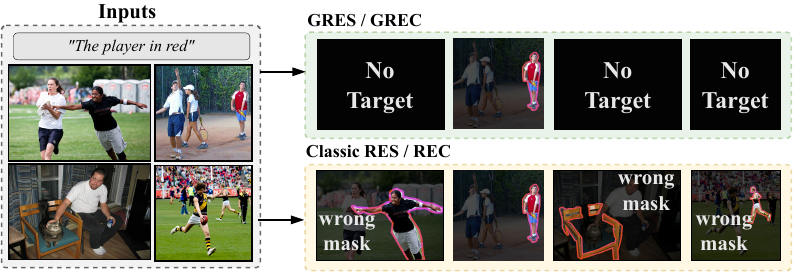}
      \vspace{-5mm}
      \caption{No-target: retrieving images that contain the object.}
      \label{fig:app_nt}
      \vspace{2mm}
  \end{subfigure} \\
  \begin{subfigure}[t]{\linewidth}
      \centering
      \includegraphics[width=\textwidth]{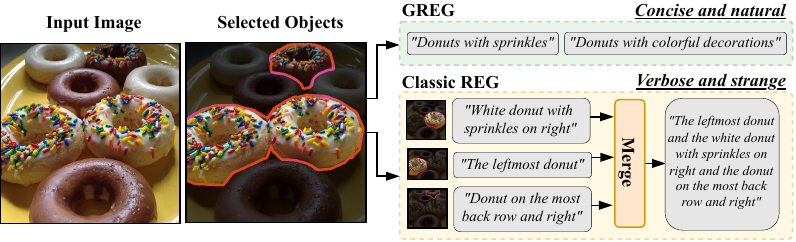}
      \vspace{-5mm}
      \caption{{GREG: capturing the common semantics and generating concise and natural expression for multiple selected objects at once.}}
      \label{fig:app_greg}
      \vspace{1mm}
  \end{subfigure}
 \vspace{-3mm}
  \caption{More applications of GREx brought by supporting multi-target and no-target expressions.}
  \label{fig:gres_app}
  \vspace{-2.16mm}
\end{figure}

\textbf{Introducing Generalized RES.} To address these limitations within classic RES, we introduce a novel benchmark termed Generalized Referring Expression Segmentation (GRES), designed to accommodate expressions indicating an unrestricted number of target objects. A GRES data sample comprises four key components: an image $I$, a language expression $T$, a ground-truth segmentation mask $M_\mathit{GT}$ that encompasses the pixels associated with all the target objects indicated by $T$, and a binary no-target label $E_\mathit{GT}$ which signifies whether the expression $T$ is devoid of any target in the image $I$. The count of objects referred to within the expression $T$ is unconstrained. GRES models take both $I$ and $T$ as inputs and produce a predicted segmentation mask, denoted as $M$. For no-target expressions, the predicted segmentation mask $M$ should entirely consist of negative, \ie, background.

\textbf{Introducing Generalized REC.} Parallel to GRES, we introduce a new benchmark called Generalized Referring Expression Comprehension (GREC), expanding from the classic REC task. In contrast to classic REC that generates a single bounding box for a sentence, GREC pursues the generation of a collection of bounding boxes, denoted as $B=\{b^i\}$, wherein each bounding box $b^i\in\mathbb{R}^4$ encloses an object among the entirety of target objects indicated by the given expression. The number of bounding boxes may vary from 0 to multiple, depending on the given expression. If the expression does not refer to any object in the image then no bounding box should be predicted.

{\textbf{Introducing Generalized REG.} GREG is a generative task that can be seen as an inverse GREC. Given an image $I$ and a set of bounding boxes $B=\{b^i\}$ or a mask $M$ defining a set of target objects in the image, the goal of GREG is to generate an expression that uniquely and unambiguously points to the set of target objects. Only one expression should be generated despite the number of target objects. In classic REG, the number of input bounding box is limited to 1, \ie, generate an expression for one object at a time.}

{\textbf{Benefits of Generalized Expressions.}} The incorporation of multi-target and no-target expressions extends the application scope beyond single object and makes the tasks more practical to real-world scenarios. This expansion facilitates the grounding of multiple targets and the exclusion of expressions that fails to indicate any target.
For example, as shown in \figurename~\ref{fig:app_mt}, the inclusion of multi-target expressions permits the utilization of phrases like ``\textit{two players on left}'' and ``\textit{all people}'' as input. This enables the selection of multiple target objects within a single inference operation. Similarly, expressions such as ``\textit{foreground}'' and ``\textit{kids}'' can be employed to achieve user-defined open vocabulary perception. This broadening of expressive possibilities significantly expands the potential applications of the tasks.
Allowing no-target expressions offers users the ability to apply the same expression to a set of images and identify which images contain the object(s) mentioned in the expression, as shown in \figurename~\ref{fig:app_nt}. This functionality proves useful when users need to locate and segment specific elements within a group of images, providing a more specific and flexible alternative to image retrieval. Additionally, the inclusion of multi-target and no-target expressions enhances the model's reliability and robustness in handling real-world scenarios where various types of expressions may occur. For example, users may unintentionally or intentionally make typographical errors in their sentences.
{Moreover, GREG enables holistic reasoning over user-selected objects to generate concise, unambiguous, and natural expressions that capture shared semantics. In contrast, classic REG typically generates one expression per object, resulting in inefficiency and a failure to capture shared attributes or relational cues, often producing redundant or awkward descriptions. For example, as shown in \figurename~\ref{fig:app_greg}, while classic REG describes each donut separately, GREG captures their shared attribute, \eg, sprinkles, with a concise and natural sentence.}

\vspace{-2.6mm}
\subsection{Evaluation Metrics for GRES} \label{sec:metrics}
\vspace{-1.96mm}

To promote diversity in GRES, we do not enforce instance-level differentiation, although our gRefCOCO dataset offers such annotations. This flexibility allows existing popular one-stage methods to be included in GRES. Besides the commonly used cumulative Intersection over Union (cIoU) and Precision@X (Pr@X), we introduce a new metric called generalized IoU (gIoU). This metric extends the mean IoU to all samples, even those without target object. 
In addition, we evaluate no-target performance using No-target-accuracy (\ntacc) and Target-accuracy (\tacc). 
 
\textbf{cIoU and Pr@X}. The cIoU metric is computed as the ratio of the total intersection pixels to the total union pixels between predicted and ground-truth foreground pixels, serving as a measure of spatial alignment between predicted and ground-truth regions. Precision@X (Pr@X) is employed in assessing the percentage of samples with IoU surpassing the predefined threshold X. As for a no-target sample, it is regarded as true positive for Pr@X if there is no predicted foreground pixel otherwise false positive.

\textbf{gIoU.} cIoU has inherent bias towards larger objects~\cite{yang2021lavt,wu2020phrasecut,GRES}. In GRES, where multi-target samples are characterized by more extensive foreground areas, this bias becomes pronounced. In response, we introduce generalized IoU~(gIoU) to rectify this inherent bias by treating all samples with equitable consideration. Similar to mean IoU, gIoU calculates the mean value of per-image IoU over all samples. For no-target expressions, the conventional per-image IoU calculation encounters a challenge that the absence of foreground pixels in the ground truth mask precludes meaningful computation. To address this challenge, gIoU adopts an approach where IoU values for true positive no-target samples are designated as 1, while the IoU values for false negative samples are assigned a value of 0.

\textbf{\ntacc\ and \tacc} assesses the model's capability in identifying no-target samples. For a no-target sample, prediction without any foreground pixels is true positive (TP), whereas a prediction with foreground pixels is false negative (FN). Then, \ntacc~(No-target accuracy) evaluates the model's performance in correctly identifying no-target samples: \ntacc~= $\frac{\mathit{TP}}{\mathit{TP}+\mathit{FN}}$. In parallel, the extent to which the model's generalization to no-target samples influences its performance on samples containing targets is measured by Target accuracy (\tacc). This metric quantifies the proportion of samples that do contain targets and are accurately classified as having targets, regardless of the correctness of the predicted segmentation mask. \tacc~= $\frac{\mathit{TN}}{\mathit{TN}+\mathit{FP}}$, where $\mathit{TN}$ represents samples with targets that are correctly identified as having targets and $\mathit{FP}$ represents samples with targets that are incorrectly identified as having no targets.

\vspace{-2.6mm}
\subsection{Evaluation Metrics for GREC} \label{sec:metrics_GREC}
\vspace{-1.96mm}

The GREC task requires generating precise bounding boxes for each individual instance of the referred targets within an image. In essence, GREC methods should exhibit the capacity to effectively differentiate between different instances. This requirement holds significance and is crucial for GREC, given that the desired outputs are bounding boxes. It ensures that the achieved outcomes align closely with the intended objective. Otherwise, there's a risk of yielding erroneous outcomes, such as predicting a single oversized bounding box that covers the entire image.

Each sample in classic REC has only one ground truth bounding box and one predicted bounding box, thus the prediction can be regarded as either a true positive (TP) or a false positive (FP). Previous classic REC methods adopt Precision@(IoU$\ge$0.5), \textit{a.k.a} top-1 accuracy, as the metric, where a prediction is considered TP if its IoU with ground truth bounding box is greater than 0.5. However, since a GREC sample has an unlimited number of ground truth bounding boxes and an unlimited number of predicted bounding boxes, the way of determining TP by IoU does not reflect the quality of prediction. To address this issue, we set a new metric for GREC: Precision@(F$_1$=1, IoU$\ge$0.5). 

\textbf{Precision@(F$_1$=1, IoU$\ge$0.5)} computes the percentage of samples that have the F$_1$ score of 1 with the IoU threshold set to 0.5, abbreviated as Pr@F$_1$. Given a sample, \ie, one expression, one image, and the predicted/ground-truth bounding boxes, a predicted bounding box is counted as a TP if it has a matched (IoU$\geq$0.5) ground-truth bounding box. If multiple predicted bounding boxes match a single ground-truth bounding box, only the one with the highest IoU is considered a TP, while the rest are treated as FP. The ground-truth bounding boxes with no matched predictions are counted as FN, while the predicted bounding boxes with no matched ground-truth are regarded as FP. We define a successful prediction of a sample as having neither FP nor FN, which leads to the maximum value 1 of F$_1$ score. As for no-target samples, the F$_1$ score is regarded as 1 if there is no predicted bounding box otherwise 0. The metric then computes the ratio of successfully predicted samples over all samples, denoted as Precision@(F$_1$=1, IoU$\ge$0.5). It is worth noting that when the all samples being evaluated consist solely of single-target expressions, the values of Precision@(F$_1$=1, IoU$\ge$0.5) and Precision@(IoU$\ge$0.5), which is used in classic REC, are equivalent.

\textbf{\ntacc\ and \tacc} For a no-target sample in GREC, prediction without any bounding box is considered a true positive (TP), otherwise false negative (FN). Then, the same as defined in GRES: \ntacc~= $\frac{\mathit{TP}}{\mathit{TP}+\mathit{FN}}$, \tacc~= $\frac{\mathit{TN}}{\mathit{TN}+\mathit{FP}}$.

\begin{figure*}[t]
  \begin{center}
     \includegraphics[width=0.996\linewidth]{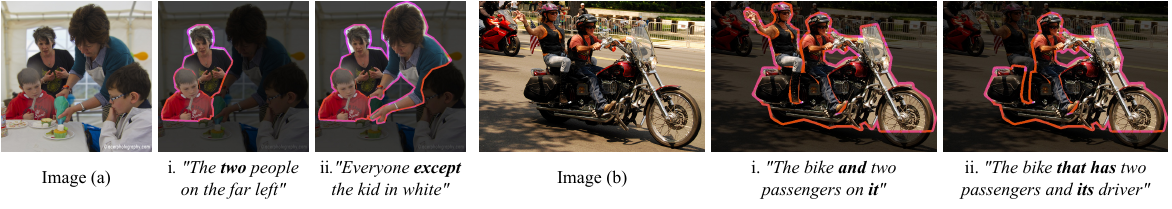}
  \end{center}
 \vspace{-6mm}
  \caption{Examples of the proposed gRefCOCO dataset.}
  \vspace{-3mm}
  \label{fig:dataset_example}
\end{figure*}

\vspace{-2.6mm}
\subsection{{Evaluation Metrics for GREG}}
\label{Sec:GREG_metrics}
\vspace{-1.96mm}

{The goal of GREG is to generate a concise, natural, and accurate expression that unambiguously captures an arbitrary set of user-selected objects with their unique or common semantics in an image. While GREG differs from classic REG in terms of input and objective, both tasks produce a single descriptive sentence. This consistency in output format allows us to evaluate GREG using the same standard metrics as REG, \ie, METEOR~\cite{banerjee2005meteor} and CIDEr~\cite{vedantam2015cider}, following prior works~\cite{sun2022proposal,tanaka2019generating,yu2017joint,liu2020attribute}.}

{\textbf{METEOR} evaluates grammatical fluency and semantic completeness by aligning candidate and reference sentences at the unigram level using exact, stem, and synonym matches. It computes a recall-weighted harmonic mean of precision and recall, with a fragmentation penalty to discourage disordered sequences. To account for the diversity in reference expressions, METEOR computes the score for each reference sentence and selects the highest among them as the final score. Higher scores indicate fluent and semantically complete expressions.}

{\textbf{CIDEr} evaluates informativeness based on consensus with a set of human-written references. Specifically, it evaluate the Term Frequency-Inverse Document Frequency (TF-IDF) weights for each $n$-gram~\cite{robertson2004understanding} between the candidate and references. This approach assigns low weight to phrases that frequently appear across the entire dataset, as they are typically uninformative, and emphasizes salient, object-specific phrases that better capture the consensus of human-written descriptions. The final score is the average cosine similarity over the 4 $n$-gram ($n$ = $1$ to $4$) levels. Higher values indicate a stronger consensus with human descriptions.}

\vspace{-2.6mm}
\subsection{gRefCOCO: A Large-scale GREx Dataset}
\vspace{-1.96mm}

To support GREx (GRES, GREC, and GREG) tasks, we construct a large-scale dataset gRefCOCO.
This dataset provides 259,859 expressions, including 90,064 multi-target expressions and 34,537 no-target expressions, referring to 61,316 distinct objects within 19,994 images. For each expression, both masks and bounding boxes of the target object(s) are provided. Additionally, a subset of single-target expressions is inherited from the RefCOCO dataset. RefCOCO, as the most widely used dataset in the field of classic REx (RES, REC, and REG), offers a wealth of high-quality single-target referring expressions. By ensuring compatibility with RefCOCO, our dataset enables seamless integration of existing REx methods into GREx tasks. This facilitates a comprehensive analysis of the performance gap of applying existing REx methods to GREx tasks. gRefCOCO dataset serves as a valuable resource for advancing research in the field of generalized referring expression.

We have developed an online annotation tool that streamlines the process of displaying images, selecting target objects, writing corresponding referring expressions, and verifying the annotated expressions. For more details about data annotation procedure and partitioning, please kindly refer to Sec.~\ref{sec:annotation}. Additionally, we conduct a comparative analysis between our newly introduced gRefCOCO dataset and RefCOCO, spotlighting the distinctive and noteworthy features of our dataset as outlined below.

{\textbf{Multi-target Samples.}} In practical scenarios, users tend to group multiple target objects in an image based on logical relationships or similarities. To account for this, annotators are given the freedom to select target instances based on their judgment instead of randomly assembling target instances. Subsequently, annotators write an unambiguous referring expression that precisely describes the selected target objects. Multi-target samples in the proposed gRefCOCO dataset exhibit 4 prominent features and challenges that deserve attention and investigation:

\noindent\textbf{1) Usage of counting expressions}, \eg, ``\textit{The \textbf{two} people on the far left}'' in \figurename~\ref{fig:dataset_example}(a). Given that RefCOCO already incorporates ordinal word numbers, \eg, ``\textit{the \textit{second} person from left}'', it becomes imperative for models to effectively distinguish between cardinal and ordinal numbers. The capability to explicitly or implicitly understand and count objects is crucial to effectively address such expressions.

\noindent\textbf{2) Compound sentence structures without geometrical relation}, such as compound sentences \textit{``A \textbf{and} B''}, \textit{``A \textbf{except} B''}, and \textit{``A \textbf{with} B \textbf{or} C''}, as shown in \figurename~\ref{fig:dataset_example}. This introduces heightened demands on models to comprehend the intricate long-range dependencies present in both the image and the sentence.

\noindent\textbf{3) Domain of attributes.}  In instances where an expression refers to multiple target objects, it is plausible for different objects to share certain attributes while also possessing distinct attributes. For example, in the phrase \textit{``the right lady in blue and kid in white''}, attributes like \textit{``right''} might be shared, whereas attributes like \textit{``blue''} and \textit{``white''} are unique to each target. This underscores the requirement for models to have a holistic grasp of all attributes and to establish meaningful connections between these attributes and their respective objects.

\noindent\textbf{4) More complex relationships}. In the context of multi-target expressions, the presence of multiple targets amplifies the frequency and intricacy of relationship descriptions, surpassing those found in single-target expressions. An illustration of this can be found in \figurename~\ref{fig:dataset_example}(b). Here, a single image hosts two distinct expressions, both employing the conjunction term \textit{``and''} and the attribute \textit{``two passengers''} for the target \textit{``bike''}. However, these two expressions point to different targets. Consequently, relationships are not only utilized to describe the nature of the target but to signify the count of targets. This requires the GREx models to possess a comprehension of all objects within the image and their interactions within the image and expression.

\textbf{No-target Samples.} 
 During the annotation process, we observed a tendency among annotators to craft numerous simplistic or generic expressions when not bound by constraints for no-target expressions. These expressions often diverged considerably from the content of other valid target-related expressions. For instance, annotators frequently generated repetitive phrases like ``\textit{dog}'' for images without any dogs present. To avoid the inclusion of such unproductive samples in the dataset, two rules are introduced for no-target expressions to enhance the diversity and difficulty:\\
 \noindent\textbf{1) The expression cannot be totally irrelevant to the image}. For example the image in \figurename~\ref{fig:dataset_example}(a), the expression ``\textit{The kid in blue}'' is permissible since there are kids present in the image, even though none of them are attired in blue. In contrast, expressions like ``\textit{airplane}'', ``\textit{tiger}'', \textit{``river''}, and so forth would be deemed unacceptable, as they bear no direct connection to any visual element within the image.\\
 \noindent\textbf{2) Annotators have the option to select a misleading expression}  from other images within the same split of RefCOCO, if it is difficult to come up with an expression that adheres to the condition mentioned in 1).

\begin{figure}[t!]
  \centering
  \hfill
  \begin{subfigure}[t]{0.49\linewidth}
      \centering
      \includegraphics[width=\textwidth]{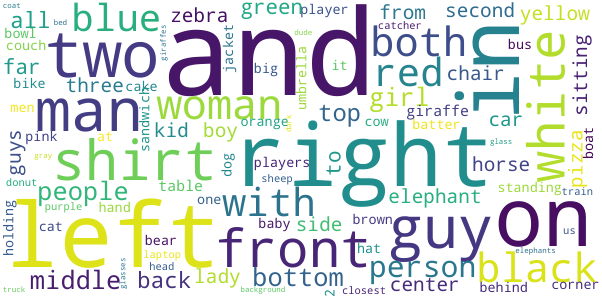} \\
      \includegraphics[width=\textwidth]{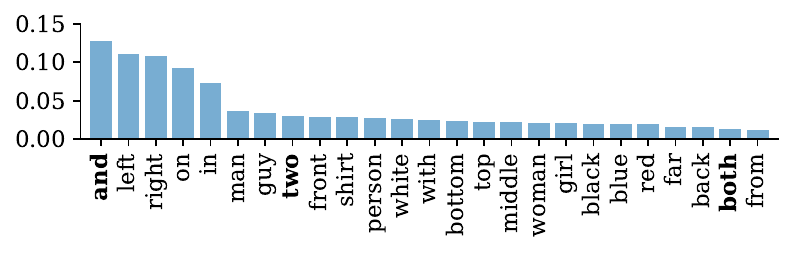} \\
      \vspace{-1mm}
      \caption{\footnotesize {Top words of gRefCOCO.}}
      \label{fig:wc_g}
  \end{subfigure}
  \hfill
  \begin{subfigure}[t]{0.49\linewidth}
      \centering
      \includegraphics[width=\textwidth]{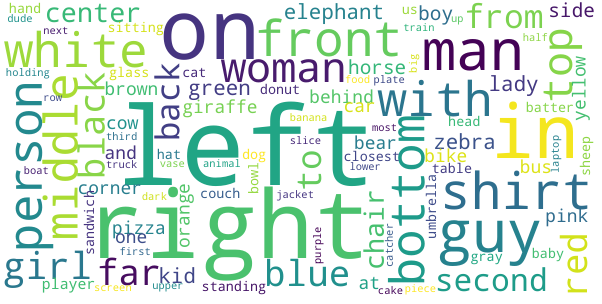} \\
      \includegraphics[width=\textwidth]{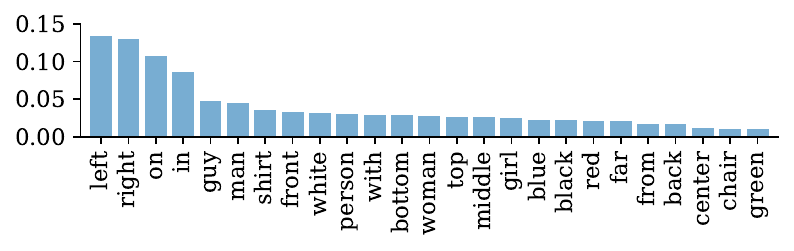} \\
      \vspace{-1mm}
      \caption{\footnotesize {Top words of RefCOCO.}}
      \label{fig:wc_r}
  \end{subfigure}
  \hfill
  \vspace{-1mm}
  \caption{Word clouds (top 100 words) and normalized~frequency histograms (top 25 words) for expressions in gRefCOCO and RefCOCO.}
  \label{fig:wc}
  \vspace{-1mm}
\end{figure}

\textbf{Word clouds} showcasing the vocabulary of the newly introduced gRefCOCO dataset and the original RefCOCO dataset are in \figurename~\ref{fig:wc_g} and \figurename~\ref{fig:wc_r}, respectively. From these figures, we can see that there are certain shared attributes between gRefCOCO and RefCOCO. Both datasets contain a significant number of words denoting relationships, such as \textit{``in''}, along with numerous attribute terms like \textit{``blue''}. Nevertheless, compared to RefCOCO, gRefCOCO exhibits some distinct traits. One of the most pronounced terms in gRefCOCO is \textit{``and''}, corresponding to the ``compound sentence structures''. Furthermore, terms related to counting, such as \textit{``two''} and \textit{``both''}, exhibit significantly greater frequency in gRefCOCO in comparison to RefCOCO.

{As we complete gRefCOCO dataset with referring expressions, segmentation masks, and bounding boxes, it can be applied to broader areas. We have already observed works that use our dataset for tasks beyond GREx. For example, GSVA~\cite{xia2024gsva} trains Multi-modal Large Language Models (MLLMs) capable of handling complex prompts and outputting masks with the help of our dataset. InstructDiffusion~\cite{geng2023instructdiffusion} uses our dataset to train diffusion-based generative image editing models that comprehend instructive prompts involving multiple instances. Our dataset also serves as a robust performance indicator for zero-shot prediction in generalist MLLMs~\cite{zhang2024psalm}. These applications further demonstrate the extensive potential uses of gRefCOCO.}

\vspace{-2mm}
\subsection{Dataset Annotation Procedure and Partitioning}\label{sec:annotation}
\vspace{-1.6mm}

\begin{figure}[t]
  \centering
  \begin{subfigure}[t]{0.45\linewidth}
      \centering
      \includegraphics[height=3cm]{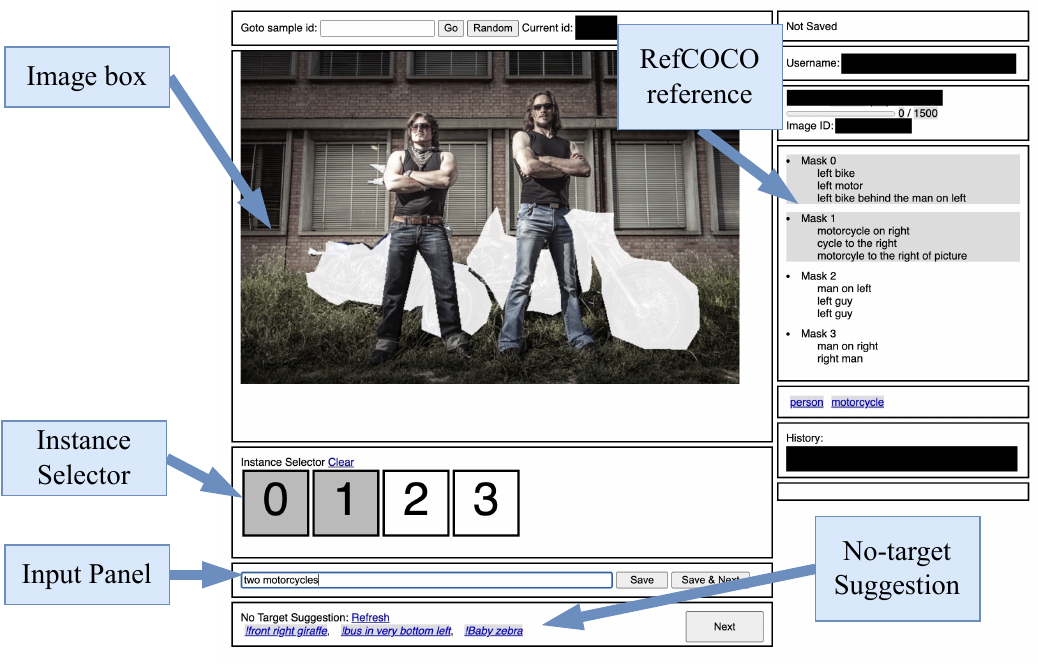}
      \caption{Annotation tool}
      \label{fig:anno}
  \end{subfigure}%
  \hfill%
  \begin{subfigure}[t]{0.4\linewidth}
      \centering
      \includegraphics[height=3cm]{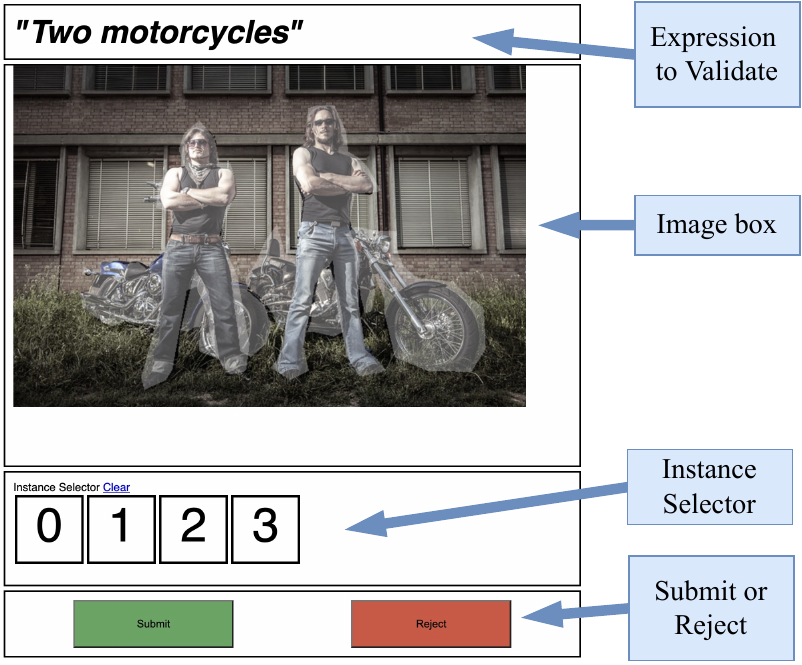}
      \caption{Validation tool}
      \label{fig:valid}
  \end{subfigure}%
  \vspace{-3.16mm}
  \caption{The screenshots of the developed annotation system used for building gRefCOCO. (Kindly zoom in).}
  \label{fig:tool}
  \vspace{-3mm}
\end{figure}

\begin{figure}[t]
  \begin{center}
     \includegraphics[width=0.92\linewidth]{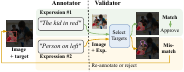}
  \end{center}
  \vspace{-5mm}
  \caption{{Interactive annotation process for gRefCOCO.}}
  \label{fig:annotation}
  \vspace{-2mm}
\end{figure}

In line with ReferIt \cite{kazemzadeh-etal-2014-referitgame}, the construction of gRefCOCO dataset follows an interactive game-like manner where annotations and validations are performed collaboratively by two players: an annotator and a validator. To streamline the annotation and validation process, we have developed a web-based annotation system comprising two components: an annotation tool for annotators and a validation tool for validators. Screenshots of the annotation system are presented in \figurename~\ref{fig:tool}. {A flowchart in Fig.~\ref{fig:annotation} illustrates the annotation process. Firstly, annotator is asked to provide a referring expression, given a target object in an image. Then, the validator is asked to find the target objects given only the image and the referring expression without knowing the ground-truth target object. If the validator can find the target object, the annotation is considered correct. Otherwise, the annotator is asked to provide a new referring expression.} This interactive annotation approach ensures the precision and quality of the annotations.

\textbf{Annotation Process.} In \figurename~\ref{fig:anno}, the annotation tool randomly selects an image from COCO dataset~\cite{lin2014microsoft} and displays all object masks in the Image Box. An annotator selects a set of targets using the Instance Selector and writes the referring expression in the Input Panel. {To help annotators write fluent and semantically rich expressions more efficiently, we use the expressions of individual objects in RefCOCO as inspirational references during gRefCOCO annotation.} After submission, the annotated sample is automatically sent for validation. The annotation system generates no-target expression suggestions by randomly selecting expressions from other images. Annotators can write no-target expressions by themselves or select deceptive expressions from the provided suggestions.

\begin{figure*}[t]
  \begin{center}
     \includegraphics[width=0.9996\linewidth]{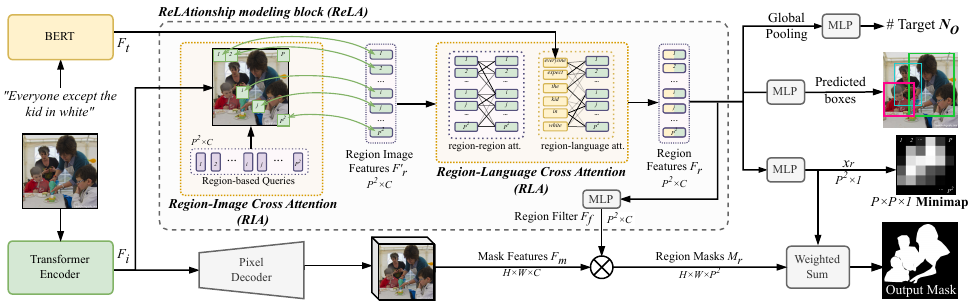}
  \end{center}
 \vspace{-5.16mm}
  \caption{Overview of the proposed baseline \textbf{ReLA}. Firstly, the given image and expression are encoded into vision feature $F_i$ and language feature $F_t$, respectively. $F_i$ is fed into a pixel decoder to produce mask features $F_m$. \textbf{ReLA}tionship modeling block takes both $F_i$ and $F_t$ as inputs and output 1) region filter $F_f$ that produces region masks $M_r$, 2) region probability map $x_r$, and 3) {number of the target objects $N_O$.}
  Output mask is obtained by weighted fusion of region masks $M_r$.}
  \vspace{-2mm}
  \label{fig:net_arch}
\end{figure*}

\textbf{Validation Process.} In \figurename~\ref{fig:valid}, the validation tool serves to validate samples received from the annotation side. The validator is presented with the image and the expression on the top of the page, and is required to independently select and submit the referred targets. The validator cannot see the annotator's selected targets and needs to find them on their own. If the targets selected by the validator match those submitted by the annotator, the sample is deemed valid. Otherwise, it is sent to another validator for a second check. Samples that still do not pass the validation are discarded. Validators can also reject samples that do not meet quality requirements or are inappropriate. For no-target samples, the validator needs to submit without instance selection and reject expressions that are not relevant to the image.

\textbf{Dataset Partitioning.} gRefCOCO follows the UNC splitting of RefCOCO~\cite{yu2016modeling} and have four non-overlapped subsets: \textit{train}, \textit{val}, \textit{testA}, \textit{testB}. The \textit{train} set is a superset of the \textit{train} set of RefCOCO, with new images added from the MS COCO training set. The images for validation and testing (\textit{val}, \textit{testA}, and \textit{testB}) are strictly identical to RefCOCO, to avoid the risk of data leakage. We would like to underscore that any form of training or pre-training for GREx tasks must exclude the images from the \textit{val}, \textit{testA}, and \textit{testB} sets of gRefCOCO dataset, which is essential to prevent any inadvertent leakage of information.

\section{The Proposed Baseline Method ReLA}

As previously mentioned, multi-target expressions present greater complexity in terms of relationship and attribute descriptions. In contrast to classic RES and REC tasks, GRES and GREC face a heightened difficulty and significance in accurately representing intricate interplays among image regions. Moreover, capturing detailed attributes for all objects adds to this challenge. To address this, our baseline approach involves explicit interaction among regions of the image and distinct words within the expression. This strategy enables a thorough analysis of their interdependencies.

Lately, many vision transformer studies, such as ViT \cite{dosovitskiy2020image}, have introduced the concept of dividing images into patches, with each patch serving as a token within the transformer. It has been observed that leveraging the attention mechanism is a convenient approach to capture the relationships between these tokens. However, as expressions frequently describe relationships and fine-grained attributes on a sub-instance level, \eg, the color of an upper body, it becomes advantageous to adopt a more soft and flexible method for obtaining these sub-instance representations. Consequently, in contrast to prior methodologies that rigidly partition images prior to the encoder phase, we introduce the ReLAtionship modeling block (ReLA). The proposed ReLA dynamically assembles semantic-related image features during the decoder phase to construct representations for individual regions. Meantime, ReLA ensures a strong correlation between region features and the actual spatial regions within the image, offering a more flexible approach.

\vspace{-2mm}
\subsection{Architecture Overview}
\vspace{-1mm}

The architecture overview of the proposed approach ReLA is shown in \figurename~\ref{fig:net_arch}. The input image undergoes processing through a transformer encoder based on Swin \cite{liu2021swin}, resulting in the extraction of a visual feature denoted as $F_i \in \mathbb{R}^{H \times W \times C}$. Here, $H$, $W$, and $C$ represent the spatial dimensions of height and width, as well as the channel dimension, respectively. The input language expression undergoes processing using the BERT \cite{devlin2018bert}, producing a language feature denoted as $F_t \in \mathbb{R}^{N_t \times C}$. $N_t$ denotes the number of words present in the input language expression, while $C$ represents the feature channel dimension. Subsequently, the vision feature $F_i$ is fed into a pixel decoder that yields the mask feature $F_m$, which is used for predicting the segmentation mask. Simultaneously, both $F_i$ and $F_t$ are directed to our proposed \textbf{ReLA}tionship modeling block (for further elaboration, please refer to Sec.~\ref{sec:ReLA}), where they undergo semantic division into $P \times P = P^2$ regions. The primary objective of this block is to explicitly model the interactions among these regions as well as among regions and languages. It's important to note that these ``regions'' correspond to the $P \times P$ patches of the image, akin to the concept found in the Vision Transformer (ViT) architecture \cite{dosovitskiy2020image}. However, unlike previous approaches~\cite{dosovitskiy2020image,xie2021segformer,strudel2021segmenter,kim2022restr} that utilize a fixed hard-split of predefined shapes and sizes for spatial areas, the ReLA block dynamically determines the shape and sizes of these spatial areas. {Also, unlike regular unconstrained instance-query~\cite{carion2020end}, we strongly link each query to a specific region in the image.} This dynamic approach ensures a more flexible and adaptable modeling of interactions among regions, setting it apart from previous methods. The ReLA block generates two sets of features: the region feature denoted as $F_r\!=\!\{f_r^n\}_{n=1}^{P^2}$ and the region filter denoted as $F_f\!=\!\{f_f^n\}_{n=1}^{P^2}$. For each of the $P^2$ regions, its corresponding region feature $f_r^n$ is used to compute a scalar $x_r^n$, which represents the probability of that region containing the target objects.

Building upon our previous work~\cite{GRES}, we propose an enhanced ReLA model with several extensions. Specifically, a box head is incorporated to extend ReLA for GREC task, enabling bounding box prediction beyond just segmentation. A target number prediction head is introduced to better handle expressions with unknown or varying target counts. Furthermore, a multi-task joint training strategy is employed to learn GRES and GREC simultaneously in a unified framework. These enhancements allow ReLA to more comprehensively address the requirements of generalized referring tasks across both segmentation and detection paradigms.

{{\textbf{For GREC task}}, an MLP is appended to $f_r^n$ for the generation of the coordinates of boxes denoted as $b^n \in \mathbb{R}^4$. Besides, the number of target objects $N_O$ is obtained by an additional global average pooling operation on $F_r$ followed by an MLP, as shown in \figurename~\ref{fig:net_arch}. The final bounding box output is denoted as $B = \{b^n\}$.}

{\textbf{For GRES task}}, the region filter $f_f^n$ is multiplied with the mask feature $F_m$, resulting in the generation of a regional segmentation mask denoted as $M_r^n\in \mathbb{R}^{H\times W}$. This mask delineates the area within the image that the specific region corresponds to.
The predicted mask for GRES is obtained by weighted aggregation of these regional segmentation masks, \ie,
\vspace{-1mm}\begin{equation}\vspace{-3mm}
  M = \sum_n(x_r^n M_r^n).
\end{equation}

\vspace{-1mm}
{\textbf{Outputs and Multi-task Joint Training.}} The predicted mask $M$ is supervised by the ground-truth target mask $M_\mathit{GT}$. The $P\times P$ probability map $x_r$ is supervised by a ``minimap'' that is downsampled from $M_\mathit{GT}$, so that we can link each region with its corresponding patch in the image. Meantime, we take the global average of all region features $F_r$ to predict the number of target objects $N_O$. {In inference, if $N_O$ is predicted to be 0, the output mask/box will be set to empty, and the number of output boxes is determined by $N_O$. $M$, $x_r$, and $N_O$ are guided by the cross-entropy loss. The predicted box $B$ is supervised by the ground-truth box $B_\mathit{GT}$. Following the training objective of MDETR~\cite{MDETR}, we use L1 and GIoU~\cite{giou} loss for bounding box $B$. The final training loss functions are:
\begin{equation}\label{Eq:loss2}
  \mathcal{L} = \lambda_M\mathcal{L}_M + \lambda_B\mathcal{L}_B + \lambda_{x_r}\mathcal{L}_{x_r} + \lambda_{N_O}\mathcal{L}_{N_O},
\end{equation}
where $\lambda_M$, $\lambda_B$, $\lambda_{x_r}$, and $\lambda_{N_O}$ are hyper-parameters to balance the losses.}

\subsection{ReLAtionship Modeling}\label{sec:ReLA}

\begin{figure}[t]
  \centering
  \vspace{0.35em}
  \hfill
  \begin{subfigure}[b]{0.493\linewidth}
      \centering
      \includegraphics[width=\textwidth]{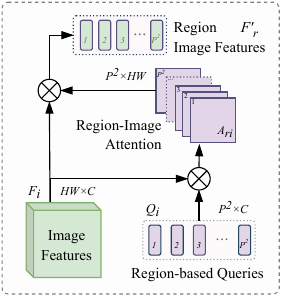}
      \caption{RIA}
      \label{fig:ria}
  \end{subfigure}
  \hfill
  \begin{subfigure}[b]{0.493\linewidth}
      \centering
      \includegraphics[width=\textwidth]{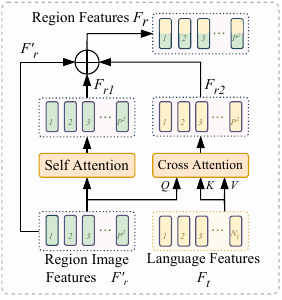}
      \caption{RLA}
      \label{fig:rla}
  \end{subfigure}
  \hfill
 \vspace{-5.6mm}
  \caption{Architectures of Region-Image Cross Attention (RIA) and Region-Language Cross Attention (RLA).}
  \label{fig:ria_rla}
\end{figure}

The proposed ReLAtionship modeling consists of two main modules: Region-Image Cross Attention (RIA) and Region-Language Cross Attention (RLA). The RIA module dynamically gathers region image features, while the RLA module focuses on capturing the relationships between regions and the language expression.

\textbf{Region-Image~Cross~Attention~(RIA).} The RIA module takes the vision feature $F_i$ and $P^2$ learnable Region-based Queries $Q_r$ as inputs. Guided by the supervision of the minimap, as shown in \figurename~\ref{fig:net_arch}, each query corresponds to a specific spatial region in the image and is tasked with decoding features for that region. The architecture of the proposed RIA module is shown in \figurename~\ref{fig:ria}. First, cross attention is conducted between the image feature $F_i$ and the $P^2$ query embeddings $Q_{r}\in \mathbb{R}^{P^2\times C}$, leading to the generation of $P^2$ attention maps:
\begin{equation}
  A_{ri} = \text{Softmax}(Q_{r}\sigma (F_{i}W_{ik})^T),
\label{eq:a-ri}
\end{equation}
where $W_{ik}\in \mathbb{R}^{C\times C}$ represents learnable parameters and $\sigma$ is GeLU~\cite{hendrycks2016gaussian}. The resulting attention maps $A_{ri}\in\mathbb{R}^{P^2\times HW}$ associate each query with a $H\times W$ attention map that indicates the relevant spatial areas in the image. The region features are then obtained from these attention maps as follows: 
\begin{equation}
F'_r=A_{ri} \sigma (F_{i}W_{iv})^T, 
\label{eq:regionFeature}
\end{equation}
where $W_{iv}\in \mathbb{R}^{C\times C}$ represents learnable parameters. This way offers more flexibility compared to rigidly dividing the image into fixed patches. Each region's feature is dynamically gathered from the corresponding relevant positions. Unlike traditional patch-based methods where each instance corresponds to a single patch, this method allows an instance to be represented by multiple regions in the minimap (as shown in \figurename~\ref{fig:net_arch}). This fine-grained region representation captures more detailed attributes at the sub-instance level, such as distinguishing the head and upper body of a person. These sub-instance representations are crucial for handling the complex relationship and attribute descriptions in GRES and GREC. $F'_r$ is fed into the RLA module to model interactions between regions and words in the expression. 
\textcolor{black}{Additionally, the region filter $F_f\in\mathbb{R}^{P^2\times C}$ obtained based on $F_r$ is utilized to predict regional segmentation masks.}

\begin{figure}[t]
  \begin{center}
     \includegraphics[width=0.996\linewidth]{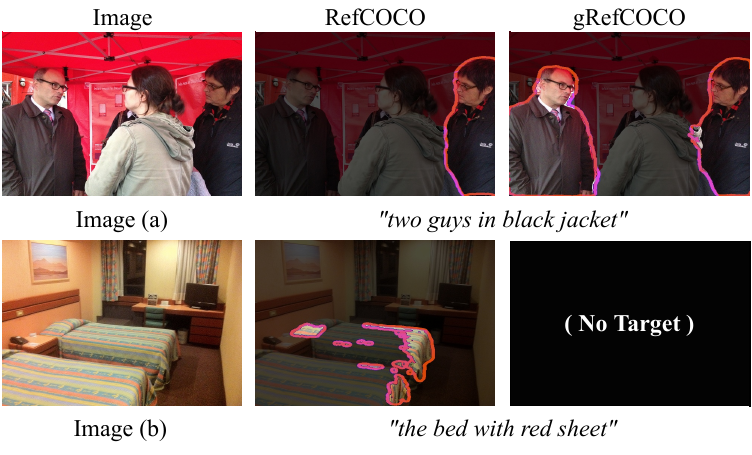}
  \end{center}
  \vspace{-5.6mm}
  \caption{Example predictions of the same model being trained on RefCOCO \textit{vs.} gRefCOCO.}
  \label{fig:gres_v_res}
 \vspace{-1.6mm}
\end{figure}

\textbf{Region-Language Cross Attention (RLA).} The region image features $F'_r$ are derived from combining image features without considering the relationships between regions and language information. To address this limitation, we introduce the RLA module, which is designed to capture interactions between regions and also interactions between regions and the language expression. As shown in \figurename~\ref{fig:rla}, the RLA module comprises a self-attention for region image features $F'_r$ and a multi-modal cross attention. The self-attention mechanism captures the dependencies between different regions. It computes the attention matrix by allowing each region feature to interact with all other regions. The resulting relationship-aware region feature is denoted as $F_{r1}$. On the other hand, the multi-modal cross attention mechanism takes the language feature $F_t$ as the Value and Key inputs, and utilizes the region image feature $F'_r$ as the Query input. This cross attention mechanism enables the model to establish relationships between regions and the linguistic content of the expression.
This multi-modal cross attention firstly models the relationship between each word and each region: 
\begin{equation}
A_{l} = \text{Softmax}(\sigma (F'_{r}W_{lq}) \sigma (F_{t}W_{lk})^T),
\end{equation}
where $A_{l}\in \mathbb{R}^{P^2\times N_t}$. Then, the RLA module generates language-aware region features denoted as $F_{r2}$ by combining the attention weights with the language feature: $F_{r2} = A_{l} F_t$. Subsequently, the interaction-aware region feature $F_{r1}$, the language-aware region feature $F_{r2}$, and the original image region features $F'{r}$ are summed together. To further integrate these three sets of features, a multi-layer perceptron (MLP) is applied, resulting in the fused region feature $F_r=\text{MLP}(F'_r+F_{r1}+F_{r2})$. $F_r$ captures the comprehensive relationships between regions, their interaction with language, and the original image features.

\section{Experiments and Discussion}

\vspace{-2mm}
\subsection{Implementation Details} \vspace{-2mm}

The proposed method uses BERT-base-uncased \cite{devlin2018bert} as language encoder. To achieve a fair comparison with previous works, single-target model utilizes Swin-base \cite{liu2021swin} backbone with feature fusing following \cite{yang2021lavt,VLTPAMI}. Images are resized to $480\times 480$ before sending into the network. The BERT language model uses the default config of huggingface's implementation~\cite{wolf-etal-2020-transformers}, and is frozen until the last two layers. The pixel decoder contains 6 Transformer decoder layers. The channel numbers of all hidden layers in the prediction head are set to 256. AdamW optimizer with a weight decay of 0.01 is used to train the whole network. Learning rate is set to 1e-5 at the beginning, and is decreased by 10 times at 11,000-th and 140,000-th iterations. {The hyper-parameters $\lambda_M$, $\lambda_B$, $\lambda_{x_r}$, and $\lambda_{N_O}$ in Eq.~(\ref{Eq:loss2}) are set to 2.0, 5.0, 0.2, and 1.0.} The model is trained for 150,000 iterations with a batch size of 48 on eight 32G V100 GPUs.

\vspace{-2mm}
\subsection{Ablation Study}
\vspace{-2mm}
\textbf{Dataset Necessity.} In order to underscore the essential nature and validity of gRefCOCO in relation to the tasks of generalized referring expression, we conduct a comparison between the outcomes of a model trained on RefCOCO and gRefCOCO. As shown in \figurename~\ref{fig:gres_v_res}, image (a) serves as a multi-target example employing a shared attribute (\textit{``in black jacket''}) to locate \textit{``two guys''}.
Despite the expression clearly indicating the presence of two target objects, the model trained on RefCOCO locates only one of them, as observed in image (a). Additionally, when presented with a no-target expression in image (b), the RefCOCO-trained model produces an inconsequential mask. These outcomes underscore the fact that models exclusively trained on single-target referring expression datasets, such as RefCOCO, lack the capacity to effectively generalize to the complexities of the GRES task.
In contrast, the newly developed gRefCOCO dataset empowers models to proficiently address expressions that refer to any arbitrary number of target objects.

\begin{table}[t]
  \renewcommand\arraystretch{1}
  \centering
  \footnotesize
 \caption{Ablation study of RIA design options.}
 \vspace{-3mm}
 \centering
     \setlength{\tabcolsep}{3.6mm}{\begin{tabular}{rl|c|cc}
      \specialrule{.1em}{.05em}{.05em} 

      &\multirow{2}{*}{Methods} & GREC & \multicolumn{2}{c}{GRES}\\
       &   & Pr@F1 & cIoU & gIoU\\
      \hline\hline
      \#1 & Hard split, input  & 53.26& 54.45 & 55.39\\
      \#2 & Hard split, decoder   & 58.19&  60.12 & 61.02\\
      \#3 & w/o minimap   & 60.07&  61.45 & 62.18 \\
      \hline
      \#4 & \textbf{ReLA} (ours)  &\textbf{61.90} & \textbf{62.91} & \textbf{63.98}\\
      \specialrule{.1em}{.05em}{.05em} 
   \end{tabular}}
  \label{tab:ablation_ria}
\vspace{-3mm}
\end{table}%

\begin{table}[t]
\renewcommand\arraystretch{1}
\centering
\footnotesize
\caption{Ablation study of RLA design options.}
\vspace{-3mm}
\centering
   \setlength{\tabcolsep}{4.mm}{\begin{tabular}{rl|c|cc}
    \specialrule{.1em}{.05em}{.05em} 
      &\multirow{2}{*}{Methods} & GREC & \multicolumn{2}{c}{GRES}\\
       &   & Pr@F1 & cIoU & gIoU\\
    \hline\hline
    \#1 & Baseline  & 56.03& 57.31 & 58.59 \\
    \#2 & + language att.  & 58.26 & 59.88 & 60.61\\
    \#3 & + region att.  & 59.86 & 61.15 & 62.48\\
    \hline
    \#4 & \textbf{ReLA} (ours)   & \textbf{61.90} & \textbf{62.91} & \textbf{63.98}\\
    \specialrule{.1em}{.05em}{.05em} 
 \end{tabular}}
\label{tab:ablation_rla}
\vspace{-3mm}
\end{table}%

\textbf{Design Options of RIA.} In \tablename~\ref{tab:ablation_ria}, we investigate the performance gain brought by RIA. In model \#1, we follow previous methods \cite{dosovitskiy2020image,kim2022restr} and rigidly split the image into $P\!\times\!P$ patches before sending them into the encoder. \tablename~\ref{tab:ablation_ria} shows that this method is not suitable for our ReLA framework, because it makes the global image information less pronounced due to compromised integrity. In model \#2, RIA is replaced by average pooling the image feature into $P\!\times\!P$. The gIoU, Pr@F1 get a significant gain of $5.63\%$, and $4.93\%$, respectively from model \#1, showing the importance of global context in visual feature encoding. Then, another $1.16\%$/$1.88\%$ gIoU/Pr@F1 gain can be obtained by adding our proposed dynamic region feature aggregation for each query (Eq.~(\ref{eq:a-ri})), showing the effectiveness of the proposed adaptive region assigning. Moreover, we study the importance of linking queries with actual image regions. In model \#3, we removed the minimap supervision so that the region-based queries $Q_r$ become plain learnable queries, resulting in a $1.80\%$/$1.83\%$ gIoU/Pr@F1 drop. This shows that explicit correspondence between queries and spatial image regions is beneficial to our model.

\textbf{Design Options of RLA.} \tablename~\ref{tab:ablation_rla} shows the importance of dependency modeling to GRES and GREC. In the baseline model \#1, RLA is replaced by point-wise multiplying region features and globally averaged language features, to achieve a basic feature fusion like previous works \cite{ding2021vision,luo2020multi}. In model \#2, the language cross attention is added onto the baseline model, bringing a gIoU/Pr@F1 gain of $2.02\%$/$2.23\%$. This shows the validity of region-word interaction modeling. Then we further add the region self-attention in model \#3 to investigate the importance of the region-region relationship, which brings a performance gain of $3.89\%$/$3.83\%$ gIoU/Pr@F1. The region-region and region-word relationship modeling together bring a significant improvement of $5.39\%$/$5.87\%$ gIoU/Pr@F1.

\begin{table}[t]
\renewcommand\arraystretch{1}
\centering
\footnotesize
\caption{Ablation study of Number of Regions.}
\vspace{-3mm}
\centering
   \setlength{\tabcolsep}{6.mm}{\begin{tabular}{c|c|cc}
    \specialrule{.1em}{.05em}{.05em} 
    \multirow{2}{*}{\#~Regions} & GREC & \multicolumn{2}{c}{GRES}\\
         & Pr@F1 & cIoU & gIoU\\
    \hline\hline
    $4\times 4$     & 55.18 & 56.64 & 57.02 \\
    $8\times 8$     & 57.62 & 59.78 & 61.30 \\
    $10\times 10$   & \textbf{61.90} & \textbf{62.91} & \textbf{63.98} \\
    $12\times 12$   & 61.04 & 62.22 & 63.71 \\
    \specialrule{.1em}{.05em}{.05em} 
 \end{tabular}}
\vspace{-3mm}
\label{tab:ablation_q}
\end{table}

\begin{figure}[t]
  \begin{center}
     \includegraphics[width=0.996\linewidth]{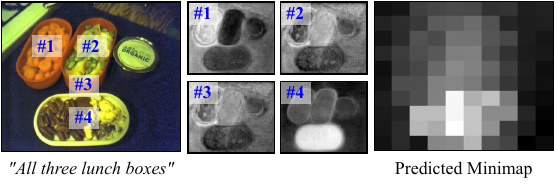}
  \end{center}
 \vspace{-5.6mm}
  \caption{The predicted region masks \& minimap.}
  \label{fig:vis_rm}
\end{figure}

\textbf{Number of Regions $P$.} 
Smaller $P$ leads to coarser regions, hindering the capture of fine-grained attributes, while larger $P$ costs more resources and decreases region area, making relationship learning more challenging. We do experiments on the selection of $P$ in \tablename~\ref{tab:ablation_q}. The model's performance improves as $P$ increases until $10$, which is selected as our setting. In \figurename~\ref{fig:vis_rm}, we visualize the predicted minimap $x_r$ and region maps $M_r$. $x_r$ displays a rough target probability of each region, showing the effectiveness of minimap supervision. We also see that the region masks capture the spatial correlation of the corresponding regions. With flexible region size and shape, each region mask contains not only the instance of this region but also other instances with strong relationships. For example, region \#4 is located inside the bottom lunch box, but as the input expression tells that all three boxes are targets, the top two also cause some responses in the output mask of region \#4.

\begin{table}[t]
\renewcommand\arraystretch{1}
\centering
\footnotesize
\caption{Ablation study of GREC output strategy.}
\vspace{-3mm}
\centering
   \setlength{\tabcolsep}{9.2pt}{\begin{tabular}{c|cccc}
    \specialrule{.1em}{.05em}{.05em} 
    Output strategy& Pr@F$_1$   & AP  & \ntacc &\tacc  \\
    \hline\hline
        Threshold    & 58.24  &  52.92  & 32.03& \ \ 99.98 \\
        Top-$k$     & 42.72 & \textbf{54.18} & \ \ 0.00 & \textbf{100.00}\\
    binary classifier   &  37.58 & 41.67& \textbf{60.29} &\ \ {97.41}   \\
    $N_O$   & \textbf{61.90}  & 53.32 & {56.37} &\ \ {96.32}   \\
    \specialrule{.1em}{.05em}{.05em} 
 \end{tabular}}
\vspace{-3mm}
\label{tab:ablation_target}
\end{table}

{
\textbf{Effect of $N_O$.} 
The GREC task requires generating a bounding box for each individual instance of the referred targets within an image. In essence, GREC methods should exhibit the capacity to effectively differentiate between different instances. The difficulty lies in how many boxes should be output. In order to control the number of outputs, we introduce a target head to predict the number of boxes that should be output within the finite set \{0, 1, 2, 3, 4, 5, 5+\}. 0 means that the current image is predicted to be a no-target output, and 5+ means that the output exceeds 5 boxes. In the 5+ case, we use a simple threshold strategy to control the output. For more detail on Threshold strategy please refer to Sec.~\ref{sec:grec_result}. We conduct experiments to verify the effectiveness of our method in \tablename~\ref{tab:ablation_target}. In the experiment, we choose hyper-parameters of both the Threshold strategy and the Top-$k$ strategy to achieve their peak performances, though this is not practical in real-world scenarios as it utilizes the ground truth information of the testing data. Our output strategy by $N_O$ is better than the best results of Threshold and Top-$k$ strategies.}

Unlike the binary flag (``target present/absent’’) used in our previous work~\cite{GRES}, the counting head $N_O$ explicitly predicts the number of target instances. This design is particularly important for GREC, which requires generating one bounding box for each of the referred target objects. In this case, any mismatch between the predicted and ground-truth number of boxes, whether more or fewer, will directly degrade performance. With $N_O$, the Pr@F1 score reaches 61.90\%, whereas replacing it with a binary classifier leads to a substantial drop to 37.58\%, as shown in \tablename~\ref{tab:ablation_target}. In contrast, GRES requires a binary mask over the entire image, without distinguishing individual instances. As a result, replacing $N_O$ in GRES has minimal effect, with cIoU and gIoU changing by less than 0.5\%. These findings demonstrate that $N_O$ plays a more crucial role when the task demands accurate instance-level predictions.

\begin{table}[t]
\renewcommand\arraystretch{1}
\centering
\footnotesize
\caption{Effect of joint training on gRefCOCO.}
\vspace{-3mm}
\centering
   \setlength{\tabcolsep}{5.mm}{\begin{tabular}{c|c|cc}
    \specialrule{.1em}{.05em}{.05em} 
    \multirow{2}{*}{Muti-task training} & GREC & \multicolumn{2}{c}{GRES}\\
    & Pr@F1& cIoU & gIoU\\
    \hline\hline
    \xmarkg    & 61.58 &62.70&63.74 \\
    \cmark     & \textbf{61.90} &\textbf{62.91}&\textbf{63.98}\\
    \specialrule{.1em}{.05em}{.05em} 
 \end{tabular}}
\vspace{-3mm}
\label{tab:ablation_j}
\end{table}

\begin{table}[t]
   \centering
      \footnotesize
   \setlength{\tabcolsep}{1.76mm}
\caption{{Computation cost analysis of the proposed method. ``Box'': MLP heads for box regression and prediction of the number of targets. Bold: best; Underline: second-best.}}
\vspace{-3mm}
{\begin{tabular}{ccc|cc|c|cc}
\specialrule{.1em}{.05em}{.05em} 
\multirow{2}{*}{RIA} & \multirow{2}{*}{RLA} & \multirow{2}{*}{Box} & \multirow{2}{*}{\#Params.}& \multirow{2}{*}{FPS} & GREC & \multicolumn{2}{c}{{GRES}}\\
 &&   &   &   & Pr@F1 & cIoU & gIoU\\
\hline\hline
\xmarkg& \xmarkg & \xmarkg & 111.6M & 21.8  & - & 50.93 &51.29    \\ 
\cmark& \xmarkg& \xmarkg  & 113.9M  &  20.3 & - & 57.24 &58.53\\ 
\xmarkg& \cmark & \xmarkg & 114.2M  & 18.9& - &  54.43 &55.34 \\ 
\xmarkg& \xmarkg & \cmark & 111.6M  & 21.8 & 49.75 & 51.04 &51.27 \\ 
\cmark& \cmark & \xmarkg & 116.6M  & 16.7 & - &   \underline{62.42} &\underline{63.60} \\ 
\cmark& \xmarkg & \cmark &  114.0M & 20.3 &\underline{56.03}& 57.31& 58.59 \\ 
\xmarkg& \cmark & \cmark &  114.2M  & 19.0 & 53.26& 54.45 &55.39\\ 
\cmark & \cmark& \cmark  & 116.7M  & 16.5 &\textbf{61.90} &\textbf{62.91} &\textbf{63.98}  \\
      \specialrule{.1em}{.05em}{.05em} 
\end{tabular}}
\label{tab:computation_resource}
\vspace{-3mm}
\end{table}

{
\textbf{Effect of Joint Training.} 
To concurrently address GRES and GREC tasks, we employ a multi-task training strategy. In \tablename~\ref{tab:ablation_j}, we conduct experiments to assess the impact of multi-task training. We can see that multi-task supervision is on par with or even slightly boosts the performance compared to the single-task variant. {It shows that our design allows GREC and GRES to benefit from each other, even when jointly trained on the same dataset without leveraging additional data, as is common in other multi-task learning settings.}
Furthermore, it validates that the RIA and RLA we designed are not only beneficial for GRES but also for GREC.
An additional advantage of our approach is the efficiency gained by training a single model to handle both GRES and GREC tasks, as opposed to dedicating one model per task. This improvement enhances the overall practicality and efficiency of the model.}

\begin{table}[t]
\renewcommand\arraystretch{1}
\centering
\footnotesize
\caption{{Ablation study of lambda parameters in Eq.~(\ref{Eq:loss2}).}}
\vspace{-3mm}
\centering
   \setlength{\tabcolsep}{2.mm}{\begin{tabular}{c|cccc|c|cc}
    \specialrule{.1em}{.05em}{.05em} 
 \multirow{2}{*}{index}   & \multirow{2}{*}{$\lambda_M$} &\multirow{2}{*}{$\lambda_B$} &\multirow{2}{*}{$\lambda_{x_r}$} &\multirow{2}{*}{$\lambda_{N_O}$} & GREC & \multicolumn{2}{c}{GRES}\\
 &  &&&      & Pr@F1 & cIoU & gIoU\\
    \hline\hline
   1& 1.0&\ \ 5.0& 0.2& 1.0    & 60.41	&60.88&	61.92 \\
   2&5.0&\ \  5.0& 0.2& 1.0      & 61.22	&62.05	&62.77 \\
   3& 2.0&\ \  2.0& 0.2& 1.0    &  60.74	&61.32	&62.45\\
   4& 2.0& 10.0& 0.2& 1.0    & 60.98	&61.70	&62.80 \\
   5& 2.0&\ \  5.0& 0.1& 1.0    &  \underline{61.89}	&62.55	&63.67\\
   6& 2.0&\ \  5.0& 0.3& 1.0    &  61.55	&62.41	&63.50\\
   7& 2.0&\ \  5.0& 0.2& 0.5    &  61.75	&\underline{62.90}	&\underline{63.81}\\
   8& 2.0&\ \  5.0& 0.2& 2.0    &  61.83	&62.68	&63.70\\
   9& \textbf{2.0}&\ \ \textbf{5.0}& \textbf{0.2}& \textbf{1.0}    &\textbf{61.90} &\textbf{62.91}&\textbf{63.98}  \\
    \specialrule{.1em}{.05em}{.05em} 
 \end{tabular}}
\vspace{-1.6mm}
\label{tab:ablation_lamabda}
\end{table}

\begin{table}[t]
  \renewcommand\arraystretch{1.05}
  \centering
  \footnotesize
  \caption{GRES results on gRefCOCO dataset.}
  \vspace{-3mm}
  \centering
     \setlength{\tabcolsep}{1.76mm}{\begin{tabular}{l|cl|cl|cl}
      \specialrule{.1em}{.05em}{.05em} 
      \multirow{2}{*}{Methods}& \multicolumn{2}{c|}{val} & \multicolumn{2}{c|}{testA} & \multicolumn{2}{c}{testB} \\
              & cIoU   & gIoU  & cIoU   & gIoU  & cIoU   & gIoU  \\
        \hline\hline
        MattNet~\cite{yu2018mattnet}   & 47.51 & 48.24 & 58.66 & 59.30 & 45.33 & 46.14  \\
        LTS~\cite{jing2021locate}   & 52.30 & 52.70 & 61.87 & 62.64 & 49.96 & 50.42 \\
        VLT~\cite{ding2021vision}   & 52.51 & 52.00 & 62.19 & 63.20 & 50.52 & 50.88 \\
        {CRIS}~\cite{wang2022cris}  & 55.34 & 56.27 & 63.82 & 63.42 & 51.04 & 51.79 \\
        LAVT~\cite{yang2021lavt}  & 57.64 & 58.40 & 65.32 & 65.90 & 55.04 & 55.83 \\
        \hline
        VLT+ReLA  & 58.65 & 59.43 & 66.60 & 65.35 & 56.22 & 57.36 \\
        LAVT+ReLA  & 61.23 & 61.32 & 67.54 & 66.40 & 58.24 & 59.83 \\
        \hline
        \cellcolor{lightgray!10}\textbf{ReLA} (ours)  & \cellcolor{lightgray!10}\textbf{62.91} &\cellcolor{lightgray!10}\textbf{63.98} &\cellcolor{lightgray!10} \textbf{69.43} & \cellcolor{lightgray!10}\textbf{70.12} &\cellcolor{lightgray!10} \textbf{60.15} &\cellcolor{lightgray!10}\textbf{61.29} \\

        \specialrule{.1em}{.05em}{.05em} 
\end{tabular}}
\vspace{-1.6mm}
  \label{tab:results_gres}
\end{table}

{\textbf{Model size and run-time speed.} In \tablename~\ref{tab:computation_resource}, we analyze the number of parameters and the time complexity for each key component. These experiments are conducted on a single NVIDIA V100 GPU based on Swin-Base~\cite{liu2021swin} backbone using the PyTorch toolkit. Our findings indicate that the proposed modules enhance performance with only a modest increase in both time and parameter complexity. {The final configuration (RIA + RLA + Box), representing our full model, achieves the best performance on both GREC and GRES tasks, while incurring only marginal overhead in parameters and inference speed.} This demonstrates that ReLA maintains a compact and efficient design, confirming its practical applicability.}

{\textbf{$\mathbf{\lambda}$ in Eq.~(\ref{Eq:loss2}).} We vary each loss weight individually while keeping the others fixed at their default values: $\lambda_M$, $\lambda_B$, $\lambda_{x_r}$, and $\lambda_{N_O}$ as 2.0, 5.0, 0.2, and 1.0. As shown in \tablename~\ref{tab:ablation_lamabda}, the default setting (index 9) yields the best performance. Varying $\lambda_M$ or $\lambda_B$ leads to notable drops (up to 1.49\% Pr@F1, 2.06\% gIoU), indicating their importance. In contrast, adjusting $\lambda_{x_r}$ or $\lambda_{N_O}$ results in minimal changes ($<$0.5\%), showing robustness to these components.}

\begin{table}[t]
    \renewcommand\arraystretch{1.05}
    \centering
    \footnotesize
    \caption{GRES no-target results on gRefCOCO dataset.}
    \vspace{-3mm}
    \centering
       \setlength{\tabcolsep}{1.6mm}{\begin{tabular}{l|cc|cc|cc}
        \specialrule{.1em}{.05em}{.05em} 
        \multirow{2}{*}{Methods}& \multicolumn{2}{c|}{val} & \multicolumn{2}{c|}{testA} & \multicolumn{2}{c}{testB} \\
                & \ntacc   & \tacc  & \ntacc   & \tacc  & \ntacc   & \tacc \\
          \hline\hline
          MattNet~\cite{yu2018mattnet}   & 41.15 & 96.13 & 44.04 & 97.56 & 41.32 & 95.32 \\
          VLT~\cite{ding2021vision}   & 47.17 & 95.72 & 48.74 & 95.86 & 47.82 & 94.66 \\
          LAVT~\cite{yang2021lavt}  & 49.32 & 96.18 & 49.25 & 95.08 & 48.46 & 95.34 \\
          \hline
          \cellcolor{lightgray!10}\textbf{ReLA}-50pix  &\cellcolor{lightgray!10}49.83 &\cellcolor{lightgray!10}96.42 &\cellcolor{lightgray!10}51.28 &\cellcolor{lightgray!10}96.39 &\cellcolor{lightgray!10}49.16 &\cellcolor{lightgray!10}95.05 \\
          \cellcolor{lightgray!10}\textbf{ReLA}  &\cellcolor{lightgray!10}\textbf{56.29} &\cellcolor{lightgray!10}\textbf{96.56} &\cellcolor{lightgray!10}\textbf{58.96} & \cellcolor{lightgray!10}\textbf{97.73} &\cellcolor{lightgray!10}\textbf{58.59} &\cellcolor{lightgray!10}\textbf{95.47} \\
          \specialrule{.1em}{.05em}{.05em} 
  \end{tabular}}
  \vspace{-3mm}
\label{tab:results_gres_nt}
\end{table}

\begin{table*}[t]
    \renewcommand\arraystretch{1.12} 
         \centering
         \footnotesize
         \caption{Results on classic RES in terms of cIoU. U: UMD split. G: Google split.}
         \vspace{-3mm}
       \setlength{\tabcolsep}{2.36mm}{\begin{tabular}{l|c!{\color{gray}\vrule}c|ccc|ccc|ccc}
          \specialrule{.1em}{.05em}{.05em} 
            \multirow{2}{*}{Methods} &\multirow{2}{*}{\shortstack{Visual\\Encoder}} & \multirow{2}{*}{\shortstack{Textual\\Encoder}}&\multicolumn{3}{c|}{RefCOCO} & \multicolumn{3}{c|}{RefCOCO+} & \multicolumn{3}{c}{G-Ref} \\
              & & & val   & test A & test B & val   & test A & test B & val$_\text{(U)}$   & test$_\text{(U)}$  & val$_\text{(G)}$\\
            \hline\hline
          MCN~\cite{luo2020multi}        &Darknet53   & GRU &62.44 & 64.20 & 59.71 & 50.62 & 54.99 & 44.69 & 49.22 & 49.40 & -     \\
          CMPC+~\cite{liu2021crossTPAMI} &Deeplab-101  & LSTM &62.47 & 65.08 & 60.82 & 50.25 & 54.04 & 43.47 & -     & -     & 49.89 \\
          EFN~\cite{feng2021encoder}     &ResNet101   & GRU &62.76 & 65.69 & 59.67 & 51.50 & 55.24 & 43.01 & -     & -     & 51.93 \\
          BUSNet~\cite{yang2021bottom}   &Deeplab-101  & Self-Att &63.27 & 66.41 & 61.39 & 51.76 & 56.87 & 44.13 & -     & -     & 50.56 \\
          LTS~\cite{jing2021locate}      &Darknet53   & GRU &65.43 & 67.76 & 63.08 & 54.21 & 58.32 & 48.02 & 54.40 & 54.25 & -     \\
          {VLT} \cite{ding2021vision}    &Darknet53   & GRU & 67.52 & 70.47 & 65.24 & 56.30 & 60.98 & 50.08 & 54.96 & 57.73 & 52.02 \\
          {ReSTR}~\cite{kim2022restr}    & ViT-B      & Transformer &67.22 & 69.30 & 64.45 & 55.78 & 60.44 & 48.27 & -     & -     & 54.48 \\
          {CRIS}~\cite{wang2022cris}     & CLIP & CLIP &70.47 & 73.18 & 66.10 & 62.27 & 68.08 & 53.68 & 59.87 & 60.36     & - \\
          {LAVT}~\cite{yang2021lavt}     & Swin-B     & BERT &72.73 & {75.82} & 68.79 & 62.14 & 68.38 & 55.10 & 61.24 & 62.09 & 60.50 \\
          {{VLT+~\cite{VLTPAMI}}} &{Swin-B}& BERT&{72.96} & {75.96} & {69.60} & {63.53} & {68.43} & {56.92} & {63.49} & \textbf{66.22} & \textbf{62.80} \\
            \hline
            \cellcolor{lightgray!10}{\textbf{ReLA} (ours)}&\cellcolor{lightgray!10}{Swin-B}&\cellcolor{lightgray!10}BERT &\cellcolor{lightgray!10}\textbf{73.82} & \cellcolor{lightgray!10}\textbf{76.48} & \cellcolor{lightgray!10}\textbf{70.18} & \cellcolor{lightgray!10}\textbf{66.04} & \cellcolor{lightgray!10}\textbf{71.02} & \cellcolor{lightgray!10}\textbf{57.65} & \cellcolor{lightgray!10}\textbf{65.00} & \cellcolor{lightgray!10}{65.97} & \cellcolor{lightgray!10}{62.70} \\
            \cellcolor{lightgray!10}{\textbf{ReLA} (ours)~$_\text{mIoU}$}&\cellcolor{lightgray!10}{Swin-B}& \cellcolor{lightgray!10}BERT & \cellcolor{lightgray!10}75.61 & \cellcolor{lightgray!10}77.79 & \cellcolor{lightgray!10}72.82 & \cellcolor{lightgray!10}70.42 & \cellcolor{lightgray!10}74.83 & \cellcolor{lightgray!10}63.87 & \cellcolor{lightgray!10}68.65 & \cellcolor{lightgray!10}69.56 & \cellcolor{lightgray!10}66.89 \\
            \specialrule{.1em}{.05em}{.05em} 
         \end{tabular}}%
         \label{tab:results_res}%
         \vspace{-2mm}
\end{table*}%

\vspace{-2mm}
\subsection{Results on GRES}
\vspace{-1mm}

\textbf{Comparison with State-of-the-art RES methods.} In \tablename~\ref{tab:results_gres}, we report the results of classic RES methods on gRefCOCO. We re-implement these methods using the same backbone as our model and train them on gRefCOCO. {It is worth noting that Segmenting Anything Model (SAM)~\cite{SAM}, a very recent powerful segmentation method trained on 11 million images, has not yet released its text prompt. Consequently, we have opted not to include SAM in the benchmark results for GRES.} For one-stage networks, output masks with less than 50 positive pixels are cleared to all-negative, for better no-target identification. For the two-stage network MAttNet~\cite{yu2018mattnet}, we let the model predict a binary label for each instance that indicates whether this candidate is a target, then merge all target instances. As shown in \tablename~\ref{tab:results_gres}, these classic RES methods do not perform well on gRefCOCO that contains multi-target and no-target samples. Furthermore, to better verify the effectiveness of explicit modeling, we add our ReLA on VLT~\cite{ding2021vision} and LAVT~\cite{yang2021lavt} to replace the decoder part of them. From \tablename~\ref{tab:results_gres}, our explicit relationship modeling greatly enhances model's performance. \Eg, adding ReLA improves the cIoU performance of the LAVT by more than $4\%$ on the val set.

\begin{figure}[t]
  \begin{center}
     \includegraphics[width=0.996\linewidth]{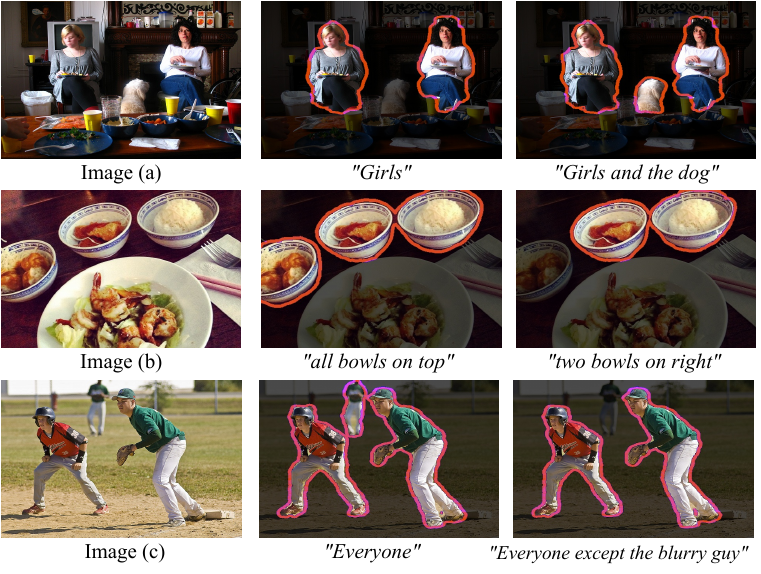}
  \end{center}
 \vspace{-5mm}
  \caption{Example results of ReLA on gRefCOCO dataset.}
  \label{fig:gRefCOCO-demo}
\vspace{-3.6mm}
\end{figure}

In \tablename~\ref{tab:results_gres_nt}, we test the no-target identification performance. In parallel with No-target-accuracy (\ntacc), the target-accuracy (T-acc.) measures the adverse effect of no-target identification on samples containing targets. As shown in the table, \tacc~of all methods are mostly higher than $95\%$, showing that our gRefCOCO does not significantly affect the model's targeting performance while being generalized to no-target samples. But from \ntacc of classic RES methods, we see that even being trained with no-target samples, it is not satisfactory to identify no-target samples solely based on the output mask. We also tested our model with the no-target classifier disabled and only use the positive pixel count in the output mask to identify no-target samples (``{ReLA}-50pix'' in \tablename~\ref{tab:results_gres_nt}). The performance is only sightly better than other methods. This shows that a dedicated no-target classifier is desired. However, although our \ntacc is higher than RES methods, there are still around $40\%$ of no-target samples are missed. We speculate that this is because many no-target expressions are very deceptive and similar with real instances in the image. We believe that no-target identification will be one of the key focuses on future research for the GRES task.

\textbf{Qualitative Results.} Some qualitative examples of our model on the val set of gRefCOCO are shown in \figurename~\ref{fig:gRefCOCO-demo}. In Image (a), our model can detect and precisely segment multiple targets of the same category (\textit{``girls''}) or different categories (\textit{``girls and the dog''}), showing the strong generalization ability. Image (b) uses counting words (\textit{``two bowls''}) and shared attributes (\textit{``on right''}) to describe a set of targets. Image (c) has a compound sentence showing that our model can understand the excluding relationship: \textit{``except the blurry guy''} and makes a good prediction.

\begin{figure}[t]
  \begin{center}
     \includegraphics[width=0.996\linewidth]{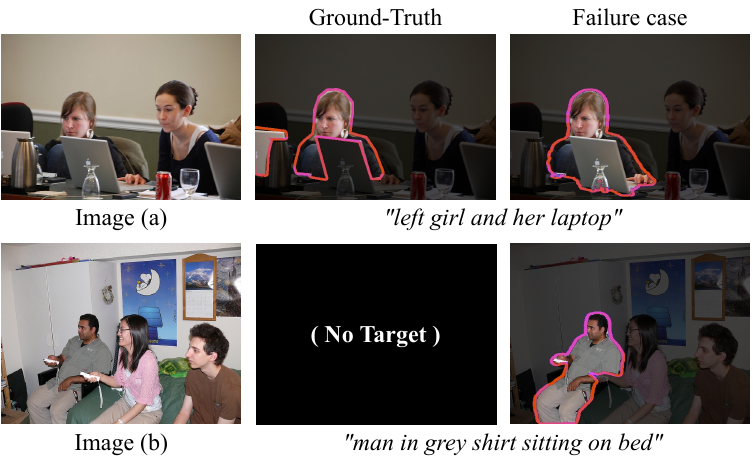}
  \end{center}
 \vspace{-3.6mm}
  \caption{GRES failure cases of ReLA on gRefCOCO dataset.}
  \label{fig:gRefCOCO-failure}
  \vspace{-3.6mm}
\end{figure}

\textbf{Failure Cases \& Discussion of GRES.} We show some failure cases of our method in \figurename~\ref{fig:gRefCOCO-failure}. Image (a) introduces a possession relationship: \textit{``left girl and \textbf{her} laptop''}. This is a very deceptive case. In the image, the laptop in center is more dominant and closer to the left girl than the left one, so the model highlighted the center laptop as \textit{``her laptop''}. Such a challenging case requires the model to have a profound understanding and comparison of all objects, and a contextual comprehension of the image and expression. In the second case, the expression is a no-target expression, referring to \textit{``man in gray shirt sitting on bed''}. In the image, there is indeed a sitting person in grey shirt, but he is sitting on a black chair very close to the bed. This further requires the model to look into the fine-grained details of all objects, and understand those details with image context.

\begin{table}[t]
      \renewcommand\arraystretch{1.16}
      \centering
           \footnotesize
     \caption{{Comparison with other methods with the same visual/textual encoders on val set of RefCOCO. All the methods are based on Swin-B~\cite{liu2021swin} and BERT~\cite{devlin2018bert}.}}
     \vspace{-3mm}
     \centering
         \setlength{\tabcolsep}{0.56mm}{\begin{tabular}{l|ccccccc}
          \specialrule{.1em}{.05em}{.05em} 
          Methods & Pr@0.5 & Pr@0.6 & Pr@0.7 & Pr@0.8 & Pr@0.9 & IoU & mIoU\\
          \hline\hline
          LTS~\cite{jing2021locate}   & 80.72 & 73.62 & 71.03 & 62.84 & 27.23 & 69.64 & 70.98 \\
          EFN~\cite{feng2021encoder}  & 82.68 & 75.00 & 72.37 & 63.26 & 29.45 & 70.83 & 72.41 \\
          LAVT~\cite{yang2021lavt}    & 84.69 & 76.82 & 75.82 & 66.58 & 34.56 & 72.63 & 74.74\\
          VLT+~\cite{VLTPAMI}   & 85.35 & 77.35 & 76.91 & 66.98 & 34.66 & 72.96 & 74.95\\
          \hline
          \rowcolor{lightgray!10}\textbf{ReLA} (ours)  & \textbf{85.92} & \textbf{83.02} & \textbf{77.71} & \textbf{68.10} & \textbf{34.99} & \textbf{73.82} & \textbf{75.61}\\
          \specialrule{.1em}{.05em}{.05em} 
       \end{tabular}}
      \label{tab:faircomp}
      \vspace{-3mm}
\end{table}%

\textbf{Results on Classic RES}. We also evaluate our method on the classic RES task and report the results in \tablename~\ref{tab:results_res}. In this experiment, our model strictly follows the setting of previous methods~\cite{ding2021vision,yang2021lavt} and is only trained on the RES datasets. As shown in \tablename~\ref{tab:results_res}, the proposed approach ReLA outperforms other methods on classic RES. Our performance is consistently higher than LAVT~\cite{yang2021lavt} with a margin of 1\%$\sim$4\% on three datasets. Although the performance gain of our proposed method over other methods on classic RES is not as significant as that on GRES, the results show that the explicit relationship modeling is not only critical to GRES but also beneficial to classic RES.

\textbf{Fair Comparison of ReLA on Classic RES.} 
To eliminate the influence of different visual/textual encoders, we compare our methods with other methods under the same visual encoder and textual encoder. In \tablename~\ref{tab:faircomp}, besides LAVT~\cite{yang2021lavt} that originally have the same backbone as ours, we re-implement three more classic RES methods: LTS~\cite{jing2021locate}, EFN~\cite{feng2021encoder}, and VLT~\cite{ding2021vision} using Swin-Base~\cite{liu2021swin} as visual encoder and BERT~\cite{devlin2018bert} as textual encoder. We test these methods on the classic RES to give a fair comparison. All methods, including ours, are trained on the RefCOCO dataset only. As shown in \tablename~\ref{tab:faircomp}, all CNN-based methods get huge performance gains with the stronger transformer-based backbones. Especially for EFN~\cite{feng2021encoder}, a performance boost of 8\% can be achieved after changing the backbone. Our method outperforms the previous state-of-the-art LAVT~\cite{yang2021lavt} by more than 1\% IoU.

\vspace{-2mm}
\subsection{Results on GREC}\label{sec:grec_result}
\vspace{-1mm}

Herein we conduct experiments under the Generalized Referring Expression Comprehension (GREC) task setting.

\begin{table}[t]
    \renewcommand\arraystretch{1.05}
    \centering
    \footnotesize
    \caption{Ablation study on Top-$k$ and Threshold strategy for multi-target and no-target samples.}
    \vspace{-2.6mm}
    \centering
       \setlength{\tabcolsep}{3.96mm}{\begin{tabular}{l|cccc}
        \specialrule{.1em}{.05em}{.05em} 
        Strategy& Pr@F$_1$   & AP  & \ntacc &\tacc  \\
        \hline\hline
        Top-1   &\ \  0.00 & 27.12 &\ \  0.00 &\textbf{100.00}\\
        Top-10  &\ \  0.00 & 53.94  &\ \  0.00  &\textbf{100.00}\\
        Top-100  &\ \  0.00 & \textbf{54.28}  &\ \  0.00  &\textbf{100.00}\\
        \hline 
        \rowcolor{lightgray!10}\textbf{Threshold}  & \textbf{52.51}  &  53.63  & \textbf{32.25}&\ \ 99.99   \\    
          \specialrule{.1em}{.05em}{.05em} 
  \end{tabular}}
  \vspace{-5mm}
\label{tab:selectingBox}
\end{table}

Existing state-of-the-art REC methods typically select the top-1 bounding box as the final output~\cite{MDETR,yu2018mattnet}, or just predict a single bounding box~\cite{luo2020multi,deng2021transvg} as the output. It is obvious that such methods cannot work for GREC task where the target objects vary from 0 to many. As shown in \tablename~\ref{tab:selectingBox}, when selecting the top-1 bounding box, the Precision@(F$_1$=1, IoU$\ge$0.5) and \ntacc~are both 0 because the top-1 strategy predicts every sample to have only one object, resulting in failures in multi-target and no-target samples. Similar problems are observed with analogous Top-$k$ strategies. Instead, our findings suggest that opting to adaptively determine output bounding boxes based on a confidence threshold is more advantageous, as shown by ``Threshold'' in \tablename~\ref{tab:selectingBox}. This approach allows the model to dynamically decide the number of bounding boxes required for each specific sample. As shown in \tablename~\ref{tab:selectingBox}, the Threshold strategy yields favorable results in terms of both Pr@(F$_1$=1, IoU$\ge$0.5) and \ntacc~metrics. 
 
\begin{table*}[t]
  \renewcommand\arraystretch{1.16}
  \begin{center}
  \footnotesize
  \caption{GREC results on gRefCOCO dataset. The original REC models have been adapted to generate multiple bounding boxes and subsequently select the target box(es) using a threshold-based criterion. Pr@F$_1$: Precision@(F$_1$=1, IoU$\ge$0.5).}\label{tab:results_grec}
  \vspace{-3mm}
     \setlength{\tabcolsep}{2.36mm}{\begin{tabular}{l|c|c|ccc|ccc|ccc}
      \specialrule{.1em}{.05em}{.05em} 
      \multirow{2}{*}{Methods} &\multirow{2}{*}{\shortstack{Visual\\Encoder}} & \multirow{2}{*}{\shortstack{Textual\\Encoder}}& \multicolumn{3}{c|}{val} & \multicolumn{3}{c|}{testA} & \multicolumn{3}{c}{testB} \\
             &  &  &Pr@F$_1$   & \ntacc  & \tacc &Pr@F$_1$   &  \ntacc  & \tacc &Pr@F$_1$    & \ntacc & \tacc  \\
        \hline\hline
        MCN~\cite{luo2020multi}   & DarkNet-53 &GRU& 28.02 &30.64 &99.62 & 32.29 & 32.04 &99.56 & 26.76 & 30.27 &99.80\\
        TransVG~\cite{deng2021transvg}  & ResNet-101 & BERT& 30.96 & 31.18 &99.50& 33.83 & 32.65 &99.50& 28.44 &32.78 &99.59  \\
        VLT~\cite{ding2021vision}   & DarkNet-53 & GRU& 36.62 & 35.20 &99.44 & 40.21 & 34.07 &99.39& 30.24 & 32.53 &99.56\\
        MDETR~\cite{MDETR}  & ResNet-101 & RoBERTa& 42.69 & 36.27 &99.40& 50.04 & 34.49 &99.99& 36.52 & 31.02 &99.63  \\
        UNINEXT~\cite{UNINEXT} & ResNet-50 & BERT&58.19  &  50.58 &96.52& 46.41 & 49.33 &96.87& 42.91 & 48.22 &98.16\\
        \hline
        \rowcolor{lightgray!10}\textbf{ReLA}~(ours) & ResNet-50 & BERT&59.36  &  55.83 &\textbf{96.36} & 48.09 & 58.73 &\textbf{98.00}& 42.85 & 57.81 &{95.44}\\

        \rowcolor{lightgray!10}\textbf{ReLA}~(ours) & Swin-B & BERT&\textbf{61.90}  &  \textbf{56.37} &{96.32}& \textbf{50.35} & \textbf{59.02} &{97.68}& \textbf{44.61} & \textbf{58.40} &\textbf{95.89}\\
        \specialrule{.1em}{.05em}{.05em} 
\end{tabular}}
\end{center}
\vspace{-6mm}
\end{table*}

Notably, the Average Precision (AP)~\!\footnote{Following COCO~\cite{lin2014microsoft}, the AP is computed by averaging over different IoU thresholds ranging from 0.50 to 0.95.} metric, which is commonly used in detection, does not be penalized too much by inclusion of numerous redundant bounding boxes characterized by low confidence scores. Consequently, a greater number of bounding boxes leads to a higher AP value. However, in the context of REC/GREC, it's imperative to avoid inundating users with an excessive number of redundant bounding boxes when they input an expression targeting at certain specific objects. Taking this into consideration, it's important to note that the AP metric doesn't well capture the performance of REC and GREC.

\begin{figure}
    \centering
    \includegraphics[width=0.86\linewidth]{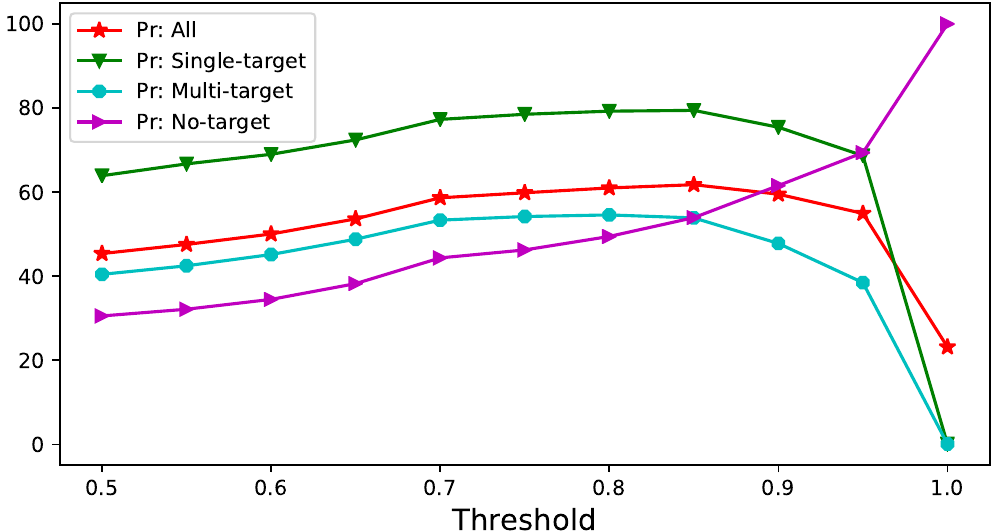}
    \vspace{-2mm}
    \caption{Effect of different threshold values for {Threshold} strategy on the performance of Pr@(F$_1$=1, IoU$\ge$0.5).}
    \label{fig:GREC-Box-Threshold}
    \vspace{-3mm}
\end{figure}

Although the Threshold strategy demonstrates acceptable performance, its performance is heavily impacted by the chosen threshold value. As shown in \figurename~\ref{fig:GREC-Box-Threshold}, raising the threshold value results in more empty output, thereby increasing the accuracy of predictions where no target present. When considering multi-target and single-target samples, their performance initially improves with an increase in the threshold value and then starts to decline. This behavior can be attributed to the fact that a higher threshold effectively filters out redundant bounding boxes, aligning with GREC's requirements. However, when the threshold becomes excessively high, some target objects may be omitted, leading to an increased number of failure cases for multi-target and single-target samples. To address this, we introduce $N_O$ to predict the number of output boxes, enhancing the practical applicability of GREC, as shown in \tablename~\ref{tab:ablation_target}.

\begin{figure}[t]
  \begin{center}
  \begin{tabular}{c@{\hspace{14mm}}c@{\hspace{16mm}}c}
  {\small Single-target} & {\small Multi-target} & {\small No-target}
  \end{tabular}
     \includegraphics[width=0.996\linewidth]{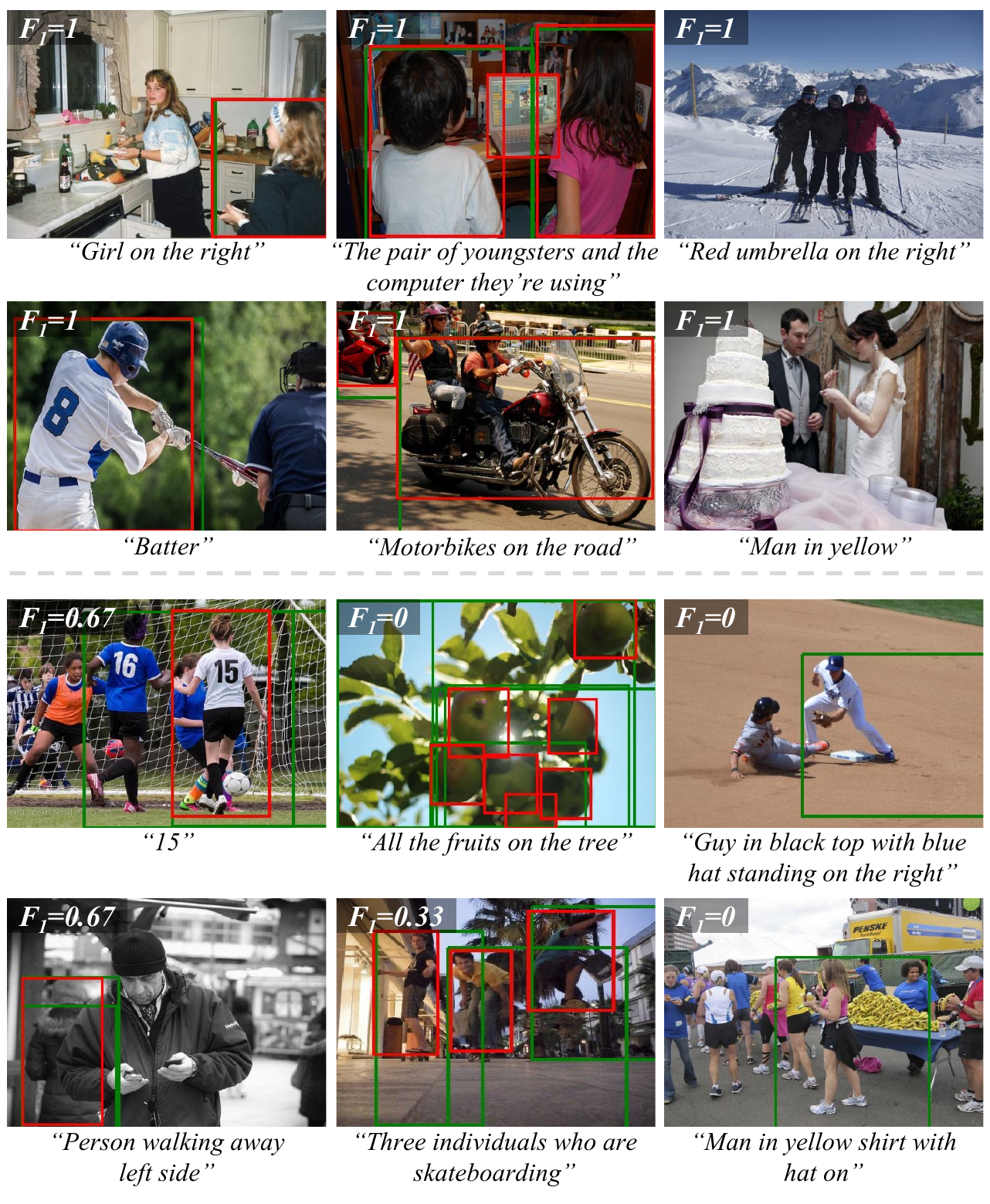}
  \end{center}
 \vspace{-5.6mm}
  \caption{Exemplary GREC results of the proposed method ReLA on gRefCOCO dataset. The ground truth and prediction are denoted by red and green bounding boxes, respectively. The first two rows showcase examples of successful outcomes, while the subsequent two rows depict examples of failure cases for single-target, multi-target, and no-target scenarios, respectively.}
  \label{fig:gRefCOCO-GREC-demo}
  \vspace{-2mm}
\end{figure}

\textbf{Qualitative Results \& Discussion of GREC.} Some qualitative examples and failure cases of the proposed method ReLA under GREC task setting are shown in \figurename~\ref{fig:gRefCOCO-GREC-demo}. The ground truth and predictive results are denoted by red bounding boxes and green bounding boxes, respectively. The first two rows demonstrate examples of successful outcomes, while the subsequent two rows show examples of failure cases for single-target, multi-target, and no-target scenarios, respectively. By analyzing the failure cases in \figurename~\ref{fig:gRefCOCO-GREC-demo}, we find that the model faces challenges when dealing with particularly misleading expressions, like \textit{``Guy in black top with blue hat standing on the right''}. This sentence presents four distinct clues, \ie, \textit{``guy''}, \textit{``black top''}, \textit{``blue hat''}, and \textit{``standing on the right''}, of which three align with elements in the image, while only \textit{``black top''} offers a clue that deviates from the objects present in the image. This example highlights the ongoing need for improving contextual understanding and nuanced interpretation of complex visual and textual cues, which is essential for enhancing its overall performance in GREC. Moreover, the extension to the multiple targets poses a box-selection challenge, particularly in the case of single-object samples. As evidenced by the two single-target failure cases in \figurename~\ref{fig:gRefCOCO-GREC-demo}, redundant bounding boxes are present, which complicates the selection process for these single-target samples. Furthermore, in the case of multi-target samples, each bounding box needs to be located accurately, otherwise it won't satisfy the requirement of IoU. For example, consider the middle image in the last row of \figurename~\ref{fig:gRefCOCO-GREC-demo}. Here, the model successfully detects three target objects and provides exactly three bounding boxes. However, only one of these bounding boxes surpasses the 0.5 IoU threshold, which ultimately leads to the failure of this particular case.

\textbf{GREC Benchmark Results on gRefCOCO.} In \tablename~\ref{tab:results_grec}, we report the benchmark results of classic REC methods and our proposed ReLA on gRefCOCO. A threshold-based criterion is employed to identify and select the final target box or boxes. Surprisingly, despite their historically impressive performance, often reaching levels exceeding 85\% in terms of Precision@(IoU$\ge$0.5), on single-target datasets like RefCOCO, the outcomes of these classic REC methods on gRefCOCO reveal a notable decline in performance. This stark contrast underscores a fundamental issue: these conventional approaches are struggling to effectively address the fresh challenges introduced by GREC. The challenges posed by GREC are multifaceted, including the need to handle multiple target objects referenced within a single expression and to resolve potential ambiguities among these objects. These complexities demand a more nuanced and advanced approach to referring expression comprehension. Consequently, these results not only highlight the limitations of conventional methods but also emphasize the urgent need for the development of more sophisticated, context-aware, and adaptable methodologies. These advanced approaches must be designed to navigate the evolving landscape of referring expression comprehension in real-world, complex scenarios. Looking ahead, the pursuit of such advanced methodologies is crucial for pushing the boundaries of the field and achieving further advancements in GREC tasks.

\subsection{{Results on GREG}}

\begin{table}[t]
      \centering
      \footnotesize
   \setlength{\tabcolsep}{0.56mm} {
      \caption{{Results of classic REG methods and MLLM methods on the proposed gRefCOCO dataset under GREG setting. $\mathcal{S}$: Single-target, $\mathcal{M}$: Multi-target.}}
      \vspace{-2.16mm}
        \resizebox{0.48\textwidth}{!}{\begin{tabular}{l|c|ccc|ccc}
      \specialrule{.1em}{.05em}{.05em} 
              & & \multicolumn{3}{c|}{METEOR} & \multicolumn{3}{c}{CIDEr} \\
              & LLM & \multicolumn{1}{c}{$\mathcal{S}$} & \multicolumn{1}{c}{$\mathcal{M}$} & \multicolumn{1}{c|}{Overall} & \multicolumn{1}{c}{$\mathcal{S}$} & \multicolumn{1}{c}{$\mathcal{M}$} & \multicolumn{1}{c}{Overall} \\
              \hline\hline
        \multicolumn{7}{l}{\textit{REG Methods}} \\
        \hline
        DisCLIP~\cite{bracha2023disclip} & \xmarkg & 10.8 &\ \  9.9 & 10.4  & 17.4  &\ \ 9.3 & 14.0\\
        Kosmos-2~\cite{peng2023kosmos}   & \cmark & 12.3 &\ \  9.2 & 10.9  & 16.0  &\ \ 5.5 & 10.9\\
        IREG~\cite{ye2023whether}        & \xmarkg & 12.9 &\ \  9.3 & 11.1 & 14.7  &\ \ 9.8 & 12.4\\
        GLaMM~\cite{GLaMM}               & \cmark & 14.0 & 10.7 & 12.5 & 18.3 & 11.9 & 15.1\\
        unleash-then-eliminate~\cite{liang2024unleashing} & \cmark & 18.6 & 14.1 & 16.9 & \textbf{22.5} & \textbf{14.8} & \textbf{18.1}\\
        \hline
        \multicolumn{7}{l}{\textit{Zero-shot MLLM-based Methods}} \\
        \hline
        GPT-4o mini~\cite{gpt4ocard}         & \cmark & 15.4  & 13.2 & 15.7  & 16.4 &\ \  9.3  & 12.2\\
        InternVL3-8B~\cite{internvl3}        & \cmark & \textbf{19.4}  & 13.5 & 17.0  & 14.0 & 10.0 & 11.6\\
        Qwen2.5-VL-7B~\cite{Qwen2.5-VL}      & \cmark & 16.3  & \textbf{14.6} & \textbf{18.1}  & 16.0 &\ \ 9.9  & 12.8\\

      \specialrule{.1em}{.05em}{.05em} 
        \end{tabular}%
        }
      \label{tab:reg_results}}%
      \vspace{-3mm}
\end{table}%

{We employ all single- and multi-target expressions in gRefCOCO for Generalized Referring Expression Generation (GREG) task. In this task, models receive an image and bounding boxes or masks of the selected objects as input and are required to generate a single expression that unambiguously refers to all selected targets. As discussed in Sec.~\ref{Sec:GREG_metrics}, we evaluate GREG using METEOR~\cite{banerjee2005meteor} and CIDEr~\cite{vedantam2015cider}.}

{\textbf{GREG Benchmark Results on {gRefCOCO}.}
\tablename~\ref{tab:reg_results} presents the results of 5 representative classic REG methods on gRefCOCO, including DisCLIP~\cite{bracha2023disclip}, Kosmos-2 \cite{peng2023kosmos}, IREG \cite{ye2023whether}, GLaMM \cite{GLaMM}, and unleash-then-eliminate \cite{liang2024unleashing}. In addition, we report results of 3 widely used Multi-modal Large Language Models (MLLMs), including the commercial closed-source model GPT-4o mini~\cite{gpt4ocard}, and the open-source models InternVL3-8B~\cite{internvl3} and Qwen2.5-VL-7B~\cite{Qwen2.5-VL}. For MLLM-based models, we adopt a zero-shot evaluation setup without any fine-tuning on the gRefCOCO dataset. Regarding the experimental setup, we overlay a transparent orange mask on the target object and feed the masked image into the model along with the following instructional prompt:
\textit{Generate a concise referring expression (within 30 words) that describes only the orange-masked object(s) in the image. Note that the mask is for indication only and not a part of the image, so do not mention the mask in your expression. Referring expressions are expressions that unambiguously describe the masked object(s) or area(s) in the image. Output in a JSON list format, \eg, [The person on the left is wearing a suit]}.}

{The results in \tablename~\ref{tab:reg_results} show that all the methods experience a marked performance drop when transitioning from single-target $\mathcal{S}$ to multi-target $\mathcal{M}$. For example, Kosmos-2~\cite{peng2023kosmos} shows a drop of 3.1 METEOR and 10.5 CIDEr on multi-target samples. Even the strongest model overall, unleash-then-eliminate \cite{liang2024unleashing}, a large language model based method specifically designed for REG, suffers a notable drop of 4.5 METEOR and 7.7 CIDEr when shifting from single-target to multi-target samples. These results underscore that referring to multiple selected objects requires more than simply scaling up single-object templates. It demands a deeper understanding of shared semantics and inter-object relationships.}

{Zero-shot MLLM-based methods~\cite{gpt4ocard,internvl3,Qwen2.5-VL} outperform classic REG methods~\cite{bracha2023disclip,peng2023kosmos,ye2023whether} on the METEOR metric, \eg, Qwen2.5-VL-7B's 18.1 \vs IREG's 11.1. This suggests that they generate more fluent and diverse sentences. However, their CIDEr scores show little improvement, \eg, Qwen2.5-VL-7B's 12.8 \vs IREG's 12.4, indicating low alignment with ground truth expressions. These methods also suffer significant performance drops from single-target to multi-target, especially in terms of CIDEr. For example, Qwen2.5-VL-7B drops by 6.1 CIDEr. These results suggest that current MLLMs, despite strong language capabilities, still struggle to ground expressions in complex multi-object visual semantics, highlighting the need for dedicated modeling of compositionality and group-level reasoning for GREG.}

\begin{figure}[t]
      \begin{center}
         \includegraphics[width=\linewidth]{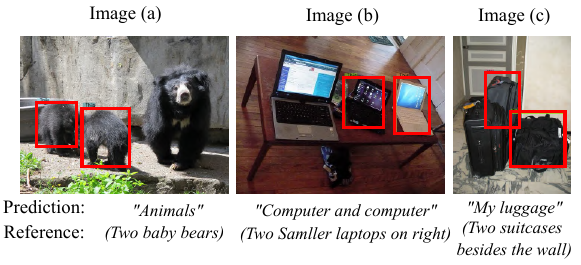}
      \end{center}
      \vspace{-5mm}
      \caption{{Failure cases of classic REG method Kosmos-2~\cite{peng2023kosmos} on multi-target expressions of gRefCOCO dataset. Prediction: predictions of Kosmos-2~\cite{peng2023kosmos}. Reference: ground truth expressions.}}
      \vspace{-2mm}
      \label{fig:reg_failure}
\end{figure}

{\textbf{Qualitative Results and Analysis.}
Failure cases in Fig.\ref{fig:reg_failure} reveal typical limitations of classic REG methods under the GREG setting. In \figurename~\ref{fig:reg_failure} (a), the model fails to understand the concept of a selected subset, mistakenly describing all three bears instead of only the two intended targets. In Fig.~\ref{fig:reg_failure} (b) and (c), the generated expressions overlook key shared attributes among the selected objects, such as ``\textit{smaller}'' or ``\textit{besides the wall}'', resulting in vague or generic descriptions that lack specificity.}

{\textbf{Discussion.} The performance degradation from single-target to multi-target cases reveals several unique challenges in GREG compared to classic REG. Specifically, the model must reason over a user-selected set of objects to generate a concise, unambiguous, and natural expression that captures shared semantics. This requires not only avoiding redundant or repetitive descriptions but also distinguishing the target subset from similar distractors based on subtle attributes, spatial layout, as well as inter-object relationships. To address these challenges, several promising solutions can be explored. First, set-aware representation learning can be employed to encode the collective semantics of the selected objects via structured aggregation or relation modeling. Second, contrastive learning between different target subsets, \eg, full versus partial selections in \figurename~\ref{fig:fig1}, can help the model capture fine-grained semantic distinctions and highlight discriminative features. Third, prompting or fine-tuning large language models (LLMs) can facilitate natural and context-aware generation. Finally, introducing multi-instance-aware decoding and leveraging synthetic data augmentation could further enhance the model’s ability to generalize to diverse GREG scenarios.}

\begin{table}[t]
\renewcommand\arraystretch{1.12} 
\centering
\footnotesize
\caption{{RVOS results on MeViS and Ref-YouTube-VOS.}}
\vspace{-3mm}
 \setlength{\tabcolsep}{0.3mm}{\begin{tabular}{r|c|ccc|ccc}
      \specialrule{.1em}{.05em}{.05em} 

&  &  \multicolumn{3}{c |}{MeViS} & \multicolumn{3}{c}{Ref-YouTube-VOS} \\
Method & Backbone & \( \mathcal{J} \)\&\( \mathcal{F} \) & \( \mathcal{J} \) & \( \mathcal{F} \)  &  \( \mathcal{J} \)\&\( \mathcal{F} \) & \( \mathcal{J} \) & \( \mathcal{F} \) \\
        \hline\hline
URVOS~\cite{seo2020urvos}& ResNet-50 &  27.8 &25.7 &29.9 & 47.2 & 45.2 &  49.1  \\
LBDT~\cite{LBDT}& ResNet-50 &  29.3&  27.8 & 30.8 & 49.3 & 48.1  &50.5  \\
MTTR~\cite{MTTR}& Video-Swin-B &   30.0 &28.8 &31.2 &58.0 & 56.8 & 59.2  \\
ReferFormer~\cite{wu2022referformer}& Video-Swin-B &  31.0 &29.8 &32.2& 62.9 & 61.3 & 64.6  \\
OnlineRefer~\cite{OnlineRefer}& Video-Swin-B &  -& -&  -&62.9& 61.0 &64.7 \\
HTML~\cite{HTML} & Video-Swin-B& -& -& -& 63.4 &61.5 &65.2 \\
VLT~\cite{VLTPAMI} & Video-Swin-B& 35.5 & 33.6&37.3 & 63.8 &61.9& 65.6\\
LMPM~\cite{mevis} & Swin-T&37.2& 34.2 &40.2 & 58.4 & 56.8 & 60.0\\
\hline
\textbf{ReLA} (ours) & Video-Swin-B&\textbf{44.6} & \textbf{41.7} &\textbf{47.5} & \textbf{65.7} &\textbf{63.8}& \textbf{67.5}\\
      \specialrule{.1em}{.05em}{.05em} 
\end{tabular}}
 \label{tab:ytvos_davis}
 \vspace{-3mm}
\end{table}

\subsection{{Results on Referring Video Object Segmentation}}

{The proposed method ReLA can also be applied to the referring video object segmentation (RVOS) task with minor adaptations. 
Firstly, we process each frame of the input video clip with our model to identify potential objects in each frame. Then based on these detected objects, we incorporate temporal modeling to capture object movements between frames, following the way of LMPM~\cite{mevis,MeViSTPAMI}.
In \tablename~\ref{tab:ytvos_davis}, we report results on the validation sets of the MeViS~\cite{mevis} and Ref-YouTube-VOS~\cite{seo2020urvos}. Ref-YouTube-VOS contains 3,978 video clips with 15,000 language expressions. MeViS is a new large-scale RVOS dataset that emphasizes more challenging motion expressions and complex scenarios, providing 28,570 language expressions for 2,006 videos of MOSE~\cite{MOSE,MOSEv2}.}

The results are evaluated using three standard metrics: region similarity ($\mathcal{J}$), contour accuracy ($\mathcal{F}$), and the mean value of the two metrics ($\mathcal{J} \& \mathcal{F} = (\mathcal{J} + \mathcal{F})/2$).
To ensure a fair comparison and maintain consistency with previous methods, we use the Video Swin Transformer~\cite{videoswin} as the backbone.
As shown in \tablename~\ref{tab:ytvos_davis}, despite ReLA not being specifically designed for video tasks, it achieves new state-of-the-art results of referring video object segmentation on both MeViS~\cite{mevis} and Ref-YouTube-VOS~\cite{seo2020urvos}. This demonstrates the effectiveness and versatility of the proposed ReLA in referring video object segmentation.

\section{Conclusion and Future Directions}

In conclusion, our study delves into the limitations of classic Referring Expression Segmentation (RES), Referring Expression Comprehension (REC), {and Referring Expression Generation (REG) tasks, highlighting their inability to handle multi-target expressions and no-target expressions. To overcome these constraints, we introduce three new GREx benchmarks: Generalized Referring Expression Segmentation (GRES), Generalized Referring Expression Comprehension (GREC), and Generalized Referring Expression Generation (GREG)}. These three benchmarks offer the flexibility to include an arbitrary number of targets in referring expressions. To support research in GREx, we build a large-scale generalized referring expression dataset named gRefCOCO. To address the GRES and GREC tasks, we present a baseline method named ReLA. This approach explicitly captures the relationships among diverse image regions and corresponding linguistic cues, resulting in remarkable performance on the newly introduced GRES/GREC tasks. The advent of GRES, GREC, and GREG relaxes the constraints on natural language or bounding box inputs, broadening the scope of application scenarios to encompass cases with multiple objects and situations where no object corresponding to the referring expression is present in the given image. This expansion paves the way for new applications like image editing/caption and beyond.

\textbf{Future Directions.} As GREx (\ie, GRES, GREC, and GREG) continues to evolve, there are several promising research directions and remaining challenges that researchers can explore. Here we provide some potential future directions for GREx. \textbf{1) Improved handling of no-target expressions and multi-target expressions.} Developing methods that better understand and identify no-target and multi-target expressions will be crucial. This involves improving the ability to distinguish between irrelevant expressions and those that contain potential references to objects, and refining models to effectively parse expressions that involve intricate relationships and attributes among multiple objects. \textbf{2) Fine-grained relationship modeling.} To handle complex expressions involving multiple objects and relationships, future works can focus on developing more advanced models for fine-grained relationship modeling. This could involve capturing more granular sub-instance level features and subtle interactions and dependencies among objects mentioned in the expressions. 3) \textbf{Robustness to noise and variation.} Real-world data often contains noise, variation, and inconsistencies. Robustness to such challenges is crucial for practical applications. Researchers can explore methods to improve the robustness of GRES and GREC models in the face of noisy or imperfect inputs. \textbf{4) Long-range dependency modeling.} GREx tasks require models to understand long-range dependencies between linguistic elements and visual context. Future research can focus on developing models that effectively capture and exploit these dependencies for more accurate prediction. \textbf{5) Handling counting and ordinal expressions.} GREx faces challenges when dealing with counting and ordinal expressions. Investigate techniques that enable models to accurately interpret and respond to expressions like \textit{``two people''} or \textit{``the second car from the left''}. \textbf{6) Cross-modal interaction and fusion.} Future research can delve deeper into the cross-modal interaction between visual and linguistic cues in GREx tasks. Exploring innovative methods for effectively fusing information from both modalities can lead to improved understanding of referring expressions.
\textbf{7) Incorporating commonsense knowledge from LLM models.} Recently, there has been a growing interest in the applications of Large Language Models (LLMs)~\cite{LLaMA,GPT-3} for dense prediction vision-language tasks~\cite{LISA,SoM,peng2023kosmos}. Integrating commonsense knowledge and reasoning capabilities from LLM can enhance the understanding of expressions that rely on implicit information or assumptions. The potential of these models to address challenges in no-target and multi-target scenarios merits further investigation. \textbf{8) Multilingual and cross-domain applications.} Expanding GREx to multilingual and cross-domain scenarios can significantly broaden their real-world applications. Developing models that can comprehend and segment referring expressions across different languages and domains is an important future direction.

{\small
    \bibliographystyle{IEEEtran}
    \bibliography{egbib}
}

\end{document}